\pdfoutput=1

\documentclass[11pt]{article}

\usepackage{array}
\newcolumntype{L}[1]{>{\raggedright\let\newline\\\arraybackslash\hspace{0pt}}m{#1}}
\newcolumntype{C}[1]{>{\centering\let\newline\\\arraybackslash\hspace{0pt}}m{#1}}
\newcolumntype{R}[1]{>{\raggedleft\let\newline\\\arraybackslash\hspace{0pt}}m{#1}}
\usepackage[preprint]{acl}

\usepackage{times}
\usepackage{latexsym}

\usepackage[T1]{fontenc}

\usepackage[utf8]{inputenc}

\usepackage{microtype}

\usepackage{inconsolata}

\usepackage{graphicx}

\usepackage{lipsum}
\usepackage{makecell}
\usepackage{booktabs}
\usepackage{multirow}

\usepackage{times}
\usepackage{latexsym}
\usepackage{amsmath}
\usepackage{amssymb}
\usepackage{physics}

\usepackage{graphicx}
\usepackage{caption}
\usepackage{subcaption}
\usepackage{hyperref}
\usepackage{enumerate}   

\usepackage{array}
\usepackage{float}
\restylefloat{table}
\usepackage{arydshln}
\usepackage{paralist}

\usepackage{mdframed}
\usepackage{enumitem}

\setlist{topsep=6pt,itemsep=4pt,partopsep=4pt, parsep=4pt}

\title{Towards Better Understanding of Program-of-Thought Reasoning \\ in Cross-Lingual and Multilingual Environments}

\author{
Patomporn Payoungkhamdee\textsuperscript{1}, 
Pume Tuchinda\textsuperscript{1},
Jinheon Baek\textsuperscript{2}
 \\
\textbf{Samuel Cahyawijaya}\textsuperscript{3},
\textbf{Can Udomcharoenchaikit}\textsuperscript{1},
\textbf{Potsawee Manakul}\textsuperscript{4}, \\
\textbf{Peerat Limkonchotiwat}\textsuperscript{5},
\textbf{Ekapol Chuangsuwanich}\textsuperscript{6}, \textbf{Sarana Nutanong}\textsuperscript{1}\\
  \textsuperscript{1}School of Information Science and Technology, VISTEC \hspace{1mm} 
  \textsuperscript{2}KAIST \hspace{1mm}
  \textsuperscript{3}Cohere \hspace{1mm} \\
  \textsuperscript{4}SCB 10X \hspace{1mm} 
  \textsuperscript{5}AI Singapore \hspace{1mm}
  \textsuperscript{6}Department of Computer Engineering,  Chulalongkorn University \\
  \texttt{\{patomporn.p\_s21, pumet\_pro,canu\_pro,snutanon\}@vistec.ac.th}
  \\
  \texttt{jinheon.baek@kaist.ac.kr, samuelcahyawijaya@cohere.com, potsawee@scb10x.com} \\
  \texttt{peerat@aisingapore.org, ekapolc@cp.eng.chula.ac.th}
}

\begin{document}
\maketitle
\begin{abstract}

Multi-step reasoning is essential for large language models (LLMs), yet multilingual performance remains challenging. 
While Chain-of-Thought (CoT) prompting improves reasoning, it struggles with non-English languages due to the entanglement of reasoning and execution. 
Program-of-Thought (PoT) prompting separates reasoning from execution, offering a promising alternative but shifting the challenge to generating programs from non-English questions. 
We propose a framework to evaluate PoT by separating multilingual reasoning from code execution to examine (i) the impact of fine-tuning on question-reasoning alignment and (ii) how reasoning quality affects answer correctness.
Our findings demonstrate that PoT fine-tuning substantially enhances multilingual reasoning, outperforming CoT fine-tuned models. 
We further demonstrate a strong correlation between reasoning quality (measured through code quality) and answer accuracy, highlighting its potential as a test-time performance improvement heuristic.

%
%
%
%
%
%

\end{abstract}




\section{Introduction}

Multi-step reasoning is crucial for large language models (LLMs), enabling them to effectively solve complex tasks, including logical, mathematical, and symbolic problems
\cite{og-cot-prompt, tot}.
Extending this capability to multilingual settings can greatly expand accessibility, allowing diverse multilingual communities to benefit from these advances. 
However, \citet{mgsm, mathoctopus} showed that LLMs perform worse in non-English languages due to differences in linguistic structure and training data. 
This finding highlights the need for approaches that tackle both multi-step reasoning and multilingual challenges.

\subsection{Research Gap}

Traditionally, multi-step reasoning has been handled through chain-of-thought (CoT)~\cite{og-cot-prompt}, which allows LLMs to tackle mathematical problem-solving by breaking down problems into sequential reasoning steps.
However, CoT requires models to handle both reasoning and computation, often leading to errors, particularly in multilingual contexts where linguistic disparity exacerbates the challenge. 
Program-of-thought (PoT) prompting~\cite{pot, pal} addresses these limitations by \emph{decoupling reasoning from computation}.
By shifting execution to an external interpreter, PoT ensures that the reasoning stage focuses solely on code generation, reducing reliance on the model's linguistic fluency in executing computational steps. 
This separation makes PoT advantageous in multilingual settings, where disparity among languages can greatly affect the model's performance.

Despite its potential, multilingualism in PoT remains underexplored.
Compared to the rich literature on non-English CoT, especially regarding multilingual fine-tuning~\cite{mathoctopus, mcot, mapo, qalign}, PoT is limited to a single cross-lingual prompting study~\cite{crosspal}.
%
This disparity emphasizes the necessity for research into PoT fine-tuning to fully exploit its potential for enhanced generalization to unseen languages and improved performance in multilingual environments.

\subsection{Problem Formulation}
\begin{figure}[htbp]
\centering
\includegraphics[width=\columnwidth]{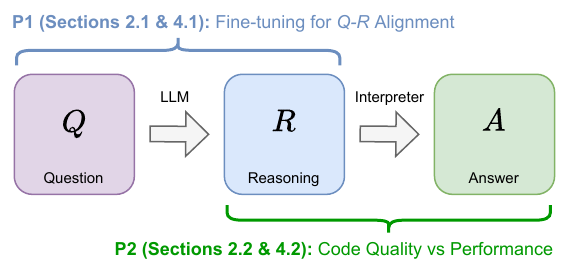}
    \caption{
    Proposed experimental framework under the PoT workflow $Q \rightarrow R \rightarrow A$. \textbf{P1}: Aligning multilingual questions (Q) with reasoning steps (R) through fine-tuning and inline comments. \textbf{P2}: Assessing the correlation between reasoning steps (R) and final answers (A) through code quality and test-time inference.
    }
    \label{fig:exp-framework}
    \vspace{-2mm}
\end{figure}

This study examines the feasibility of decoupling natural language reasoning from computation in non-English languages.
We formalize multilingual PoT within a two-stage framework as shown in Figure~\ref{fig:exp-framework}: (i) 
$Q \rightarrow R$,  where the model generates reasoning steps $R$ from questions $Q$; (ii) $R \rightarrow A$, where an external interpreter executes $R$ to obtain the final answer $A$.
Our research is organized into two problems: P1 and P2, as follows.

\noindent
\textbf{(P1) Fine-tuning for Q-R Alignment.}
This study attempts to answer the question:
\emph{
How can we \textbf{align questions $Q$} posed in different languages with effective \textbf{reasoning steps $R$} in PoT, and how do \textbf{fine-tuning decisions} influence cross-lingual and multilingual reasoning performance?
}

We evaluate different fine-tuning decisions under two fine-tuning scenarios.
\begin{compactitem}
\item \emph{Cross-lingual:} The model is fine-tuned only in English and evaluated cross-lingual zero-shot.
%
\item \emph{Multilingual:} Training data includes samples in target languages, allowing direct $Q$-$R$ alignment in target languages.
\end{compactitem}
These examinations analyze the impact of fine-tuning choices on reasoning alignment, providing insights into how language availability influences cross-lingual and multilingual PoT performance.

We explore multiple decisions regarding the use of inline comments in PoT reasoning, evaluating their impact in both cross-lingual and multilingual settings.
For the cross-lingual setting, since the model is fine-tuned only in English, we compare keeping English comments versus removing them entirely. 
Our results show that removing comments leads to better generalization in unseen languages.
For the multilingual setting, with access to target-language training data, we evaluate keeping English comments versus translating them into the target language. 
%
Our findings indicate that translating comments improves reasoning alignment, reflecting overall performance.

\noindent
\textbf{(P2) Code Quality vs Performance.}
This study attempts to answer the question: 
\emph{
To what extent does the \textbf{code quality} of reasoning steps $R$ \textbf{affect the correctness} of final answers $A$, and how can we use this knowledge to \emph{improve PoT performance}?
}

We investigate the relationship between PoT performance and code quality measured through \texttt{ICE-Score}~\cite{ice-score}, which quantifies the correctness of intermediate steps within the code.
Our analysis reveals a strong correlation between them.
Building on this insight, we employ the ICE score as a heuristic for test-time scaling within the Soft Self-Consistency~\cite{soft-sc} method, implementing it as a form of soft voting.
Experimental results show that this simple adjustment outperforms the standard Self-Consistency~\cite{wang2023selfconsistency} baseline, where models generate multiple candidates and apply hard voting.
In particular, this approach improves the overall accuracy across languages, in cross-lingual settings, increasing the performance from 31.6\% to 56.6\%.

\subsection{Contributions}

The contributions of our work are as follows:
\begin{compactitem}
    
\item \textbf{Experimental Framework for Multilingual PoT --- }
We exploit the reasoning-execution disentanglement in PoT to break down the problem into two key challenges: $Q$-$R$ alignment (how multilingual questions map to reasoning steps) and $R$-$A$ association (how reasoning quality translates into correct answers).

\item \textbf{Systematic Evaluation of Fine-Tuning for $Q$-$R$ Alignment ---} We investigate how fine-tuning impacts multilingual PoT performance under cross-lingual and multilingual settings, analyzing the role of inline comments.

\item \textbf{Correlation Between Code Quality and Answer Accuracy ---} 
We assess how the quality of generated reasoning steps $R$ influences the correctness of the final answers $A$ and leverage this insight to improve test-time inference. 
\end{compactitem}

\section{Proposed Studies}

\subsection{Fine-tuning for Q-R Alignment (P1)}

\begin{figure}[!t]
    \includegraphics[width=\linewidth, trim={0 0 0 0.1cm}, clip]{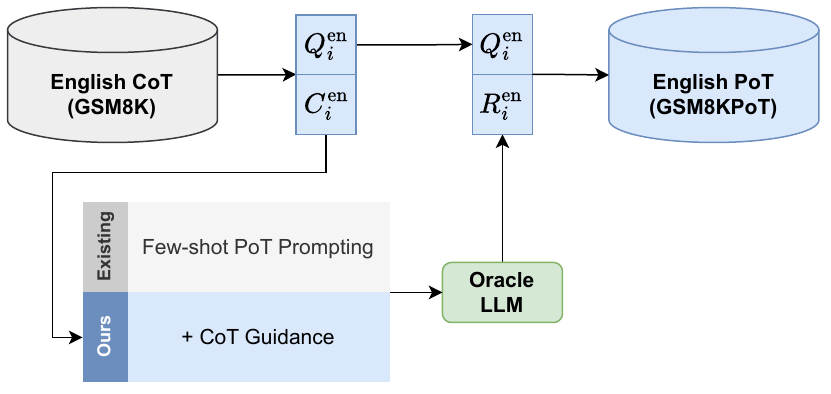}
    \caption{
    The generation pipeline for \texttt{GSM8KPoT}, in which a PoT answer ($\vb*{R}_i^\text{en}$) is synthesized using the Oracle LLM, with additional natural language reasoning ($\vb*{C}_i^\text{en}$) provided as guidance.
    }
    \label{fig:pipeline-cross}
\end{figure}

%
To fairly compare PoT and CoT, we use the \emph{Grade School Math} (\texttt{GSM8K}) dataset \cite{cobbe2021gsm8k} and explore three prompting strategies for generating PoT with an Oracle LLM: (i) zero-shot PoT, (ii) few-shot PoT, and (iii) the proposed few-shot PoT with CoT guidance, 
as shown in Figure~\ref{fig:pipeline-cross}.
%
Zero-shot PoT generates Python solutions without examples. Few-shot PoT improves this with two solved examples while adding CoT guidance further enhances program generation.
This structured guidance substantially improves accuracy, achieving a 96.1\% correctness rate in PoT-generated samples and leading to the development of the \texttt{GSM8KPoT} dataset (details in Appendix~\ref{ap:pot-syn}).

Examining PoT in multilingual settings is challenging due to the scarcity of datasets that align questions across multiple languages with structured reasoning steps.
To address this, we construct dataset variants (outlined in Table~\ref{tab:compare-metods}) to evaluate cross-lingual and multilingual fine-tuning.
%
We control for language effects by varying the languages of questions and inline comments, allowing us to assess the impact of each fine-tuning strategy.

In the context of cross-lingual and multilingual PoT, inline comments can potentially play a crucial role. As established by \citet{mgsm}, the language used in multi-step reasoning processes, such as those in CoT reasoning, is a key design consideration.
We hypothesize that the design choices for inline comments in PoT function similarly to the language considerations in CoT. 
Thus, we analyze its implications comprehensively.

%


\subsubsection{Cross-lingual Setup}
The cross-lingual setup comprises two datasets: one with inline comments in the reasoning steps and one without, defined as follows.
\begin{table}[!t]
  \vspace{-2mm}

  \centering
  \renewcommand{\arraystretch}{1.4}

  \resizebox{\columnwidth}{!}{
  \begin{tabular}{C{1.0cm}|C{1cm}|C{1cm}|C{2.5cm}|C{0.3cm}}
    \hline
    \textbf{Setup} & Lang. of $Q$ & Lang. of Comm. in $R$ & Dataset & Eq. \\
    \hline
    Cross & En & En & $\mathcal{D}^\texttt{GSM8KPoT}_\texttt{en}$  & \ref{eq:gsm8kpot}\\
    \cdashline{2-5}
    & En & \emph{nc}  &  $\mathcal{D}^\texttt{GSM8KPoT}_\texttt{nc}$ & \ref{eq:gsm8kpot-nc} \\
    \hline
    Multi & En & Multi & $\mathcal{D}^\texttt{MGSM8KPoT}_\texttt{cross-comment}$ & \ref{eq:mgsm8kpot-cross-comment}\\
     \cdashline{2-5}
     & Multi & En & $\mathcal{D}^\texttt{MGSM8KPoT}_\texttt{cross-question}$ & \ref{eq:mgsm8kpot-cross-question}\\    

    \cdashline{2-5}
     & Multi & Multi & $\mathcal{D}^\texttt{MGSM8KPoT}_\texttt{parallel}$ & \ref{eq:mgsm8kpot-parallel} \\

    \cdashline{2-5}
     & Multi & \emph{nc} & $\mathcal{D}^\texttt{MGSM8KPoT}_\texttt{nc}$ & \ref{eq:mgsm8kpot-nc} \\
    \hline
  \end{tabular}
  }
  \vspace{-2mm}
  \caption{
  Our proposed study employs multiple approaches, leveraging the question-comment characteristics within the dataset to compare different best fine-tuning strategies. \emph{NC} stands for ``no comment''.
  }
  \label{tab:compare-metods}
\end{table}

\vspace{2mm}
\noindent
\underline{\emph{En-En}} --- 
%
%
We employ \texttt{GSM8KPoT} as the foundational dataset, which can be formally represented by the following equation.
\begin{equation}
\mathcal{D}^\texttt{GSM8KPoT}_\texttt{en} = \{(\vb*{Q}_i^\text{en}, \vb*{R}_i^\text{en})\}_{i=1}^N,
\label{eq:gsm8kpot}
\end{equation}
where the questions $\vb*{Q}_i^\text{en}$ are obtained from English GSM8K, and the synthesized intermediate reasoning in the programming language ($\vb*{R}_i^\text{en}$) include inline comments in English. Note that the superscript \text{en} in $\vb*{R}_i^\text{en}$ denotes the language of code comments. 

\vspace{2mm}
\noindent
\underline{\emph{En-nc}} --- We also include a variant with all comments removed.
\begin{equation}
\mathcal{D}^\texttt{GSM8KPoT}_\texttt{nc} = \{(\vb*{Q}_i^\text{en}, \vb*{R}_i^\text{nc})\}_{i=1}^N,
\label{eq:gsm8kpot-nc}
\end{equation}
where the superscript ``$\text{nc}$'' in the reasoning steps $\vb*{R}_i^\text{nc}$ stands for ``no comment''.

%
%
%


\subsubsection{Multilingual Setup}


The multilingual setup comprises four datasets.
Following the concept proposed in MGSM8KInstruct~\cite{mathoctopus}, we consider CoT cross and CoT parallel strategies, varying how the languages of questions and inline comments match or mismatch.
%
%
%
We also include a no-comment variant, pairing multilingual questions with reasoning steps that exclude comments.
These four datasets are defined as follows. 

\vspace{2mm}
\noindent
\underline{\emph{En-Multi}} --- Following the CoT definition in CoT Cross, we translate English inline comments using machine translation (\texttt{MT}), producing program reasoning in target languages:
    \begin{equation}
        \mathcal{D}^\texttt{MGSM8KPoT}_\texttt{cross-comment} = \{(\vb*{Q}_i^\text{en}, \vb*{R}_i^l)|l\in L_\text{all}\}_{i=1}^N,
        \label{eq:mgsm8kpot-cross-comment}
    \end{equation}
where $L_\text{all}$ denotes the language set. 
The superscript $l$ in $\vb*{R}_i^l$ is a variable representing a language. 

\vspace{2mm}
\noindent
\underline{\emph{Multi-En}} --- This variant provides multilingual questions $\vb*{Q}_i^l$ by applying machine translation to $\vb*{Q}_i^\text{en}$, while keeping the inline comments in English.
    \begin{equation}
        \mathcal{D}^\texttt{MGSM8KPoT}_\texttt{cross-question} = \{(\vb*{Q}_i^l, \vb*{R}_i^\text{en})|l\in L_\text{all}\}_{i=1}^N.
        \label{eq:mgsm8kpot-cross-question}
    \end{equation}

\vspace{2mm}
\noindent    
\underline{\emph{Multi-Multi}} --- Both questions and inline comments are in the same language $l$:
    \begin{equation}
        \mathcal{D}^\texttt{MGSM8KPoT}_\texttt{parallel} = \{(\vb*{Q}_i^l, \vb*{R}_i^l)|l\in L_\text{all}\}_{i=1}^N.
        \label{eq:mgsm8kpot-parallel}
    \end{equation}    
Note that in this case, the superscript $l$ in $\vb*{Q}_i^l$ and $\vb*{R}_i^l$ denotes the fact that both question and inline comments are in the same language. 

\vspace{2mm}
\noindent
\underline{\emph{Multi-nc}} --- Similar to $\mathcal{D}^\texttt{GSM8KPoT}_\texttt{nc}$, we also include a no-comment variant for this setup. 
    \begin{equation}
        \mathcal{D}^\texttt{MGSM8KPoT}_\texttt{nc} = \{(\vb*{Q}_i^l, \vb*{R}_i^\text{nc})|l\in L_\text{all}\}_{i=1}^N.
        \label{eq:mgsm8kpot-nc}
    \end{equation}





\subsection{Code Quality Analysis (P2)} \label{subsec:code_analysis}


%
%
After addressing the multilingual problem with PoT, the task is split into two parts: multi-step reasoning via code generation and execution via a Python interpreter for numerical computations.
While the interpreter ensures arithmetic accuracy, the challenge lies in generating syntactically and logically correct programs free of errors.

%
%
We assess code quality using the \texttt{ICE-Score} \cite{ice-score}, which measures usefulness (how well the code addresses the query) and functional correctness (evaluated through intermediate validation with an Oracle LLM). Our focus is functional correctness, rating program validity from 0 (incorrect/incomplete) to 4 (fully correct).

%
We use \texttt{ICE-Score} to assess whether improved alignment strategies enhance both accuracy and code quality. 
Furthermore, to compare code quality with final answer accuracy, we conduct two analyses:
\begin{inparaenum}[(i)]
    \item \emph{System level}: Spearman correlation assesses whether higher-quality code improves overall model performance.
    \item \emph{Sample level}: AUC and t-test assess whether code validity can determine answer correctness.
\end{inparaenum}

%
\textbf{Test-time Scaling.} We investigate the use of \texttt{ICE-Score} to enhance model inference in test-time scaling. Building on Self-Consistency (\texttt{SC}) \cite{wang2023selfconsistency}, which generates multiple reasoning candidates and applies majority voting (hard voting), we extend this approach with Soft Self-Consistency (\texttt{Soft-SC}) \cite{soft-sc}.
\texttt{Soft-SC} refines this process by averaging the \texttt{ICE-Score} for each final answer candidate, ranking responses by overall code quality. This shift from hard to soft voting may improve performance.

\subsection{Discussions}

The six datasets enable us to examine how language alignment and inline comments impact cross-lingual and multilingual PoT reasoning. 
Inline comments act as alignment anchors between questions and reasoning steps expressed in a programming language. 
However, they can hinder cross-lingual generalization to unseen languages. 
In this respect, we aim to understand (i) how multilingual data availability influences PoT’s ability to generate accurate reasoning steps and (ii) how inline comments affect performance across language setups.

Code quality analysis provides an intermediate observation linking these decisions to the accuracy of the final answer. 
By examining both aspects, we establish a structured understanding of how multilingual data and inference-time strategies interact to improve PoT performance, laying the groundwork for our experimental validation in Section~\ref{section:code-analysis-results}.

\begin{table*}[ht]
\tiny
  \centering
  \resizebox{\textwidth}{!}{
  \begin{tabular}{l|llllllllll|l}
    \hline
    \textbf{Method} & en & de & fr & es & ru & zh & ja & th & sw & bn & All \\
    \hline
    \multicolumn{1}{l|}{\underline{Llama2-7B}} &     
    \multicolumn{1}{c}{} & \multicolumn{6}{c}{} & \multicolumn{3}{c|}{} & \multicolumn{1}{c}{} \\    
    CoT & 43.6 & 32.4 & 30.4 & 30.4 & 26.4 & 25.2 & 15.2 & 4.8 & 2.0 & 5.6 & 21.6 \\
    PoT & \textbf{58.0} & \textbf{40.4} & \textbf{40.4} & \textbf{43.6} & \textbf{37.1} & \textbf{38.4} & \textbf{32.7} & \textbf{7.6} & \textbf{5.6} & \textbf{12.0} & \textbf{31.6} \\
    \hline
    \multicolumn{1}{l|}{\underline{CodeLlama 7B}} & 
    \multicolumn{1}{c}{} & \multicolumn{6}{c}{} & \multicolumn{3}{c|}{} & \multicolumn{1}{c}{} \\  
    CoT & 43.2 & 33.2 & 32.8 & 39.6 & 26.8 & 27.2 & 18.8 & 16.4 & 3.2 & 9.2 & 25.0 \\
    PoT & \textbf{58.8} & \textbf{48.4} & \textbf{51.6} & \textbf{53.6} & \textbf{49.8} & \textbf{41.6} & \textbf{39.6} & \textbf{26.8} & \textbf{4.4} & \textbf{11.2} & \textbf{38.6} \\
    \hline    
    \multicolumn{1}{l|}{\underline{Llama2-13B}} &     
    \multicolumn{1}{c}{} & \multicolumn{6}{c}{} & \multicolumn{3}{c|}{} & \multicolumn{1}{c}{} \\     
    CoT & 47.4 & 39.2 & 37.6 & 41.2 & 38.0 & 35.2 & 18.8 & 7.2 & \textbf{7.4} & 6.8 & 27.9 \\
    PoT & \textbf{64.0} & \textbf{52.4} & \textbf{54.4} & \textbf{55.6} & \textbf{51.2} & \textbf{44.0} & \textbf{40.0} & \textbf{13.9} & 7.2 & \textbf{13.6} & \textbf{39.6} \\
    \hline        
    \multicolumn{1}{l|}{\underline{Llama3-8B}} &    
    \multicolumn{1}{c}{} & \multicolumn{6}{c}{} & \multicolumn{3}{c|}{} & \multicolumn{1}{c}{} \\ 
    CoT & 62.8 & 51.2 & 52.8 & 54.8 & 45.2 & 40.0 & 33.6 & 39.6 & 28.0 & 39.6 & 44.8 \\
    PoT & \textbf{68.4} & \textbf{62.2} & \textbf{59.2} & \textbf{62.4} & \textbf{60.4} & \textbf{52.4} & \textbf{45.4} & \textbf{43.6} & \textbf{34.8} & \textbf{46.0} & \textbf{53.5} \\    
    \hline
  \end{tabular}
  }
  \caption{
  Accuracy (\%) on MGSM in \textbf{cross-lingual setting}.
  }
  \label{tab:main-cross}

\end{table*}

\section{Experimental Setup}

\textbf{Base LLMs.} We conduct experiments with various base LLMs, using Llama2-7B \cite{Llama2} as the foundation for the following variants:
\begin{compactenum}[i)]
\item \emph{Code-specific variant}: CodeLlama-7B \cite{codeLlama}, optimized for code and programming-related tasks.
\item \emph{Size variant}: Llama2-13B \cite{Llama2}, a larger-scale version of Llama2.
\item \emph{Version variant}: Llama3 8B \cite{Llama3}, a more recent iteration with enhanced multilingual capabilities. 
\end{compactenum}

\noindent\textbf{Oracle LLM.} To ensure reproducibility, we employ Llama3.1-405B Instruct \cite{Llama3} as our Oracle model for generating the PoT dataset and assessing the quality of the code.


\vspace{2mm}
\noindent \textbf{Evaluation.}
We evaluate model performance by measuring accuracy on the MGSM \cite{mgsm} dataset in a zero-shot setting using greedy decoding.
The study includes the following languages: English (en), German (de), French (fr), Spanish (es), Russian (ru), Chinese (zh), Japanese (ja), Thai (th), Swahili (sw), and Bengali (bn).
For CoT evaluation, numerical outputs are extracted via regular expressions and compared to labels, following \citet{mathoctopus}.
For PoT evaluation, generated programs are executed in a Python interpreter, with outputs compared to labels for accuracy.

\vspace{2mm}
\noindent \textbf{Measures}: As outlined in Table~\ref{tab:compare-metods}, in cross-lingual setting, we finetune each LLM independently on GSM8K \cite{cobbe2021gsm8k} and GSM8KPoT, using both \(\mathcal{D}^\texttt{GSM8KPoT}_\texttt{en}\) and \(\mathcal{D}^\texttt{GSM8KPoT}_\texttt{nc}\) variants.
For multilingual CoT, we finetune each LLM separately on MGSM8K Instruct Parallel and Cross \cite{mathoctopus}.
For multilingual PoT, we utilize the generated answers from GSM8KPoT and map the questions for each language in MGSM8K Instruct to create MGSM8KPoT.
To study the effects of inline comments, we create versions of GSM8K and MGSM8KPoT without inline comments by removing them from the original datasets.
Additionally, we generate a variation of MGSM8KPoT by applying machine translation.
We utilize nllb-200-distilled-600M~\cite{nllb} for translating inline comments, ensuring coverage across all languages in this study.

\section{Experimental Results}


\subsection{Impact of Q-R Alignment Fine-tuning}
\label{sec:QR-Alignment}
\subsubsection{Cross-lingual Setting}
The experimental results presented in Table \ref{tab:main-cross} indicate that PoT consistently outperforms CoT across all languages and model classes, achieving superior results in 39 out of 40 cases. The only exception is Swahili in the Llama2-13B model, where PoT reached an accuracy of 7.2\%, compared to CoT's 7.4\%, showing only a slight difference.

When comparing models of the same size, CodeLlama-7B consistently outperforms Llama2-7B in most languages. 
The improvements are notable in non-English languages such as German (+8.0), French (+11.2), and Thai (+19.2), suggesting that the incorporation of code data during pretraining improves structured reasoning even in cross-lingual settings.
Scaling up to Llama2-13B leads to further improvements over both Llama2-7B and CodeLlama-7B. 
While model size remains an important factor in boosting overall accuracy, the strong performance of CodeLlama-7B relative to Llama2-7B indicates that increased code data during pretraining \cite{codeLlama} can enhance reasoning ability.
For models with enhanced multilingual capabilities, such as Llama3-8B, where the performance gap between languages is narrower, the results suggest that PoT remains more effective in cross-lingual settings, achieving superior accuracy across unseen languages.

In Table \ref{tab:pot-inline-comment}, we compare performance when fine-tuning between \(\mathcal{D}^\texttt{GSM8KPoT}_\texttt{en}\) and \(\mathcal{D}^\texttt{GSM8KPoT}_\texttt{nc}\).
Overall, training without comments tends to improve non-English accuracy across Llama2 models for both 7B and 13B variants, where omitting comments reduces English accuracy slightly but yields larger gains in non-English languages, like German and Bengali, boosting the overall score.
\begin{table}[htbp]
\tiny
  \centering
  \renewcommand{\arraystretch}{1.2}

  \resizebox{\columnwidth}{!}{
  \begin{tabular}{l|c|cc|c}
    \hline
    \textbf{Method} & en & de & bn & ALL \\
    \hline\hline
    \multicolumn{1}{l|}{\underline{Llama2-7B}}& \multicolumn{1}{c|}{} & \multicolumn{2}{c|}{} & \multicolumn{1}{c}{} \\
    With Comments & \textbf{58.3} & 37.9 & 9.9 & 30.0  \\
    Without Comments & 58.0 & \textbf{40.4} & \textbf{12.0} & \textbf{31.6}  \\   
    \hline     
    \multicolumn{1}{l|}{\underline{CodeLlama-7B}}& \multicolumn{1}{c|}{} & \multicolumn{2}{c|}{} & \multicolumn{1}{c}{} \\
    With Comments & \textbf{61.4}  & 45.2 & \textbf{15.6} & 36.6  \\
    Without Comments & 58.8  & \textbf{48.4} & 11.2 & \textbf{38.6}  \\
    \hline    
    \multicolumn{1}{l|}{\underline{Llama2-13B}}& \multicolumn{1}{c|}{} & \multicolumn{2}{c|}{} & \multicolumn{1}{c}{} \\
    With Comments & \textbf{67.3}  & 48.4 & 13.2 & 37.4  \\
    Without Comments & 64.0  & \textbf{52.4} & \textbf{13.6} & \textbf{39.6}  \\
    \hline
    \multicolumn{1}{l|}{\underline{Llama3-8B}}& \multicolumn{1}{c|}{} & \multicolumn{2}{c|}{} & \multicolumn{1}{c}{} \\
    With Comments & 46.4  & 48.2 & 37.5 & 40.6  \\
    Without Comments & \textbf{68.4}  & \textbf{62.2} & \textbf{46.0} & \textbf{53.5}  \\      
    \hline            
  \end{tabular}
  }
  \caption{The impact of code comments on accuracy across different models in cross-lingual setup. The ALL score is from Appendix~\ref{ap:full-main}.}
  \label{tab:pot-inline-comment}
\end{table}

Interestingly, CodeLlama-7B shows mixed results: including comments helps in English and Bengali, whereas excluding comments improves German and also leads to a higher overall score.
This may reflect the specialized training corpus for CodeLlama, which emphasizes code tokens and might interact differently with inline explanations.  

Finally, Llama3-8B shows the largest swing: removing comments substantially boosts performance for all languages (including English), suggesting that inline explanations can sometimes distract or misalign the Q–R.
Taken together, these findings indicate that, for most models, \(\mathcal{D}^\texttt{GSM8KPoT}_\texttt{nc}\) provides better cross-lingual generalization and more robust Q–R alignment.

\begin{table*}[htbp]
\tiny
\centering
\resizebox{\textwidth}{!}{
\begin{tabular}{l|llllllllll|l}
\hline
\textbf{Method} & en & de & fr & es & ru & zh & ja & th & sw & bn & All \\
\hline
\multicolumn{1}{l|}{\underline{Llama2-7B}} & & & & & & & & & & & \\
CoT Cross & 45.2 & 38.4 & 36.8 & 40.0 & 33.2 & \textbf{33.6} & 23.6 & 16.8 & \textbf{18.8} & 17.2 & 30.4 \\
PoT Cross Comment & \textbf{54.8} & \textbf{47.2} & \textbf{51.2} & \textbf{46.2} & \textbf{42.8} & 33.2 & \textbf{34.8} & \textbf{20.0} & 17.6 & \textbf{18.0} & \textbf{36.6} \\
\hdashline
CoT Parallel & 48.8 & 42.4 & 44.0 & 42.4 & 38.0 & 42.4 & 31.6 & 33.6 & 34.4 & 27.6 & 38.5 \\
PoT Parallel & \textbf{56.0} & \textbf{47.2} & \textbf{46.4} & \textbf{54.0} & \textbf{49.6} & \textbf{44.4} & \textbf{40.0} & \textbf{40.4} & \textbf{37.6} & \textbf{30.8} & \textbf{44.6} \\
\hline
\multicolumn{1}{l|}{\underline{CodeLlama2-7B}} & & & & & & & & & & & \\
CoT Cross & 47.6 & 38.8 & 33.2 & 38.8 & 35.2 & 31.6 & 28.8 & 23.6 & 17.2 & 20.4 & 31.5 \\
PoT Cross Comment & \textbf{58.0} & \textbf{47.2} & \textbf{51.4} & \textbf{52.4} & \textbf{48.0} & \textbf{44.2} & \textbf{38.0} & \textbf{28.8} & \textbf{20.4} & \textbf{22.4} & \textbf{41.1} \\
\hdashline
CoT Parallel & 46.0 & 40.0 & 38.8 & 44.0 & 43.2 & 41.2 & 35.6 & 41.6 & 30.8 & 32.0 & 39.3 \\
PoT Parallel & \textbf{61.9} & \textbf{52.8} & \textbf{54.4} & \textbf{52.4} & \textbf{53.6} & \textbf{50.4} & \textbf{44.8} & \textbf{44.8} & \textbf{39.6} & \textbf{35.6} & \textbf{49.0} \\
\hline
\multicolumn{1}{l|}{\underline{Llama2-13B}} & & & & & & & & & & & \\
CoT Cross & 58.4 & 50.4 & 46.4 & 49.6 & 43.6 & \textbf{43.2} & 33.6 & \textbf{25.6} & \textbf{23.6} & \textbf{24.4} & 39.9 \\
PoT Cross Comment & \textbf{62.0} & \textbf{53.6} & \textbf{52.4} & \textbf{54.8} & \textbf{50.0} & 42.0 & \textbf{39.2} & 21.6 & 23.2 & 23.2 & \textbf{42.2} \\
\hdashline
CoT Parallel & 60.8 & 53.6 & 52.0 & 54.4 & 52.8 & 53.6 & 45.2 & 43.6 & 41.2 & 38.0 & 49.5 \\
PoT Parallel & \textbf{63.5} & \textbf{56.4} & \textbf{59.2} & \textbf{59.2} & \textbf{55.2} & \textbf{54.0} & \textbf{51.6} & \textbf{50.0} & \textbf{52.8} & \textbf{44.4} & \textbf{54.6} \\
\hline
\multicolumn{1}{l|}{\underline{Llama3-8B}} & & & & & & & & & & & \\
CoT Cross & 69.2 & 58.0 & 54.8 & 58.0 & 57.2 & 50.0 & 44.4 & 40.4 & 40.4 & 42.0 & 51.4 \\
PoT Cross Comment & \textbf{72.8} & \textbf{62.4} & \textbf{66.4} & \textbf{67.2} & \textbf{63.6} & \textbf{52.0} & \textbf{49.6} & \textbf{52.0} & \textbf{46.2} & \textbf{51.2} & \textbf{58.3} \\
\hdashline
CoT Parallel & 66.8 & 53.6 & 57.2 & 60.8 & 62.4 & 60.0 & 50.8 & 57.6 & 53.6 & 54.8 & 57.8 \\
PoT Parallel & \textbf{76.5} & \textbf{64.4} & \textbf{63.2} & \textbf{66.4} & \textbf{64.0} & \textbf{63.2} & \textbf{56.4} & \textbf{57.6} & \textbf{59.6} & \textbf{55.2} & \textbf{62.6} \\
\hline
\end{tabular}
}
\caption{
Accuracy (\%) on MGSM in \textbf{multilingual setup}. 
}
\label{tab:main-multi}
\vspace{-3mm}
\end{table*}

\subsubsection{Multilingual Setting}
Table \ref{tab:main-multi} shows that PoT continues to outperform CoT in multilingual settings across all languages and model variants.
A few exceptions appear in some languages within the Llama 2 family trained on $\mathcal{D}^\texttt{MGSM8KPoT}_\texttt{cross-comment}$, yet overall results confirm that PoT provides consistent advantages over CoT.

As observed in the cross-lingual experiments, CodeLlama2-7B maintains its advantage over Llama2-7B across all languages in the multilingual setting.
This performance gap is particularly pronounced in French (+8.0), Chinese (+6.0), and English (+5.9), further suggesting that increasing code data during pretraining yields stronger reasoning capabilities.
Scaling to larger models continues to deliver gains, with Llama2-13B showing consistent improvements over both 7B variants.
However, the most dramatic improvements come from Llama3-8B, which achieves substantially higher accuracy across all languages, reaching 76.5\% in English while maintaining strong performance even in non-English languages like Thai (57.6\%) and Bengali (55.2\%).
The stronger gains highlights the benefits of explicit multilingual training over multilingual transfer, emphasizing the role of scaling and adaptation in optimizing reasoning across languages.
%

%
Finally, we investigate the most effective way to align Q and R in a multilingual context.
As shown in Table \ref{tab:multilingual-ablation}, translating inline comments into the target language consistently yields superior performance across all model variants. 
We hypothesize that this improvement comes from the enhanced semantic alignment between code and natural language when comments are presented in the target language during training.
In summary, these findings indicate that $\mathcal{D}^\texttt{MGSM8KPoT}_\texttt{parallel}$ provides the optimal Q-R alignment for multilingual settings.
\begin{table}[htbp]
\small
  \centering
  \resizebox{\columnwidth}{!}{
  \begin{tabular}{l|c|cc|c}
    \hline
    \textbf{Method} & en & de & bn & ALL \\
    \hline\hline
    \multicolumn{1}{l|}{\underline{Llama2-7B}}& \multicolumn{1}{c|}{} & \multicolumn{2}{c|}{} & \multicolumn{1}{c}{} \\
    PoT Cross Comment & \underline{54.8} & \textbf{47.2} & 18.0 & 36.6  \\
    PoT Cross Question & 46.0 & 37.6 & 28.8 & 37.7  \\ 
    PoT Parallel & \textbf{56.0} & \textbf{47.2} & \textbf{30.8} & \textbf{44.6} \\
    PoT No Comment & 53.6 & 41.6 & \underline{29.2} & \underline{40.6}  \\ 
    \hline
    \multicolumn{1}{l|}{\underline{CodeLlama2-7B}}& \multicolumn{1}{c|}{} & \multicolumn{2}{c|}{} & \multicolumn{1}{c}{} \\
    PoT Cross Comment & \underline{58.0} & 47.2 & 22.4 & 41.1  \\
    PoT Cross Question & 48.0 & 42.8 & 28.8 &  40.5 \\ 
    PoT Parallel & \textbf{61.9} & \textbf{52.8} & \textbf{35.6} & \textbf{49.0}  \\
    PoT No Comment & 56.8 & \underline{47.6} & \underline{35.2} & \underline{45.6}  \\ 
    \hline    
    \multicolumn{1}{l|}{\underline{Llama2-13B}}& \multicolumn{1}{c|}{} & \multicolumn{2}{c|}{} & \multicolumn{1}{c}{} \\
    PoT Cross Comment & \underline{62.0} & \underline{53.6} & 23.2 & 42.2  \\
    PoT Cross Question & 53.0 & 47.6 & \underline{35.9} & 45.1  \\ 
    PoT Parallel & \textbf{63.5} & \textbf{56.4} & \textbf{44.4} & \textbf{54.6}  \\
    PoT No Comment & 58.4 & 51.6 & \underline{35.2} & \underline{46.4}  \\ 
    \hline    
    \multicolumn{1}{l|}{\underline{Llama3-8B}}& \multicolumn{1}{c|}{} & \multicolumn{2}{c|}{} & \multicolumn{1}{c}{} \\
    PoT Cross Comment & \underline{72.8} & \underline{62.4} & \underline{51.2}	& \underline{58.3}  \\
    PoT Cross Question & 37.2 & 30.3 & 30.4 & 31.6  \\ 
    PoT Parallel & \textbf{76.5} & \textbf{64.4} & \textbf{55.2} & \textbf{62.6}  \\
    PoT No Comment &  65.2 & 60.0 & 48.4 & 56.5  \\ 
    \hline    
  \end{tabular}
  }
  \caption{
  The impact of various fine-tuning strategies is examined, where PoT Cross includes either comment-only or question-only translation. In contrast, the Parallel approach involves either the exclusion of comments or the inclusion of translated comments.
  }
  \label{tab:multilingual-ablation}
\end{table}

\subsection{R-A Relationship: Code Quality Analysis}
\label{section:code-analysis-results}

\textbf{Does Better Strategy Improve Code Quality?}
As discussed in Section~\ref{subsec:code_analysis}, we assess code quality across alignment strategies in cross-lingual and multilingual settings, focusing on Llama2-7B and CodeLlama-7B.
Table~\ref{tab:code-quality} shows that higher accuracy correlates with better code quality.
Additionally, code quality in lower resource languages, like Bengali, is much lower than in English and German, which aligns with the accuracy trends.
This finding reflects the inherent challenges of generating code in low-resource languages, where model performance is typically more constrained.
\begin{table}[htbp]
 \tiny
  \centering
  \resizebox{\columnwidth}{!}{
  \begin{tabular}{l|c|cc|c}
    \hline
    \textbf{Method} & en & de & bn & ALL \\
    \hline\hline
    \multicolumn{5}{c}{\textit{Cross-lingual}} \\
    \hline    
    \multicolumn{1}{l|}{\underline{Llama2-7B}}& \multicolumn{1}{c|}{} & \multicolumn{2}{c|}{} & \multicolumn{1}{c}{} \\
    With Comments & \textbf{2.49} & \textbf{1.87} & 0.45 & 1.39  \\
    Without Comments & \textbf{2.49} & \textbf{1.87} & \textbf{0.49}	& \textbf{1.44}  \\ 
    \hline   
    \multicolumn{1}{l|}{\underline{CodeLlama2-7B}}& \multicolumn{1}{c|}{} & \multicolumn{2}{c|}{} & \multicolumn{1}{c}{} \\
    With Comments & \textbf{2.66} & 2.06 & \textbf{0.61}	&1.97  \\
    Without Comments & 2.55 & \textbf{2.13} & 0.54 & \textbf{2.02}  \\
    \hline\hline
    \multicolumn{5}{c}{\textit{Multilingual}} \\
    \hline
    \multicolumn{1}{l|}{\underline{Llama2-7B}}& \multicolumn{1}{c|}{} & \multicolumn{2}{c|}{} & \multicolumn{1}{c}{} \\
    PoT Cross Comment & \underline{2.56} & 2.41 & 1.26	&1.98  \\
    PoT Cross Question & 2.32 & 2.07 & 1.52	&2.03  \\
    PoT Parallel & \textbf{2.83} & \textbf{2.55} & \textbf{1.96} & \textbf{2.45}  \\
    PoT No Comment & 2.54 & \underline{2.16} & \underline{1.71}	& \underline{2.13}  \\
    
    \hline
    \multicolumn{1}{l|}{\underline{CodeLlama2-7B}}& \multicolumn{1}{c|}{} & \multicolumn{2}{c|}{} & \multicolumn{1}{c}{} \\
    PoT Cross Comment & \underline{2.84} & 2.40 & 1.34	& 2.15  \\
    PoT Cross Question & 2.45 & 2.23 & 1.54	&2.11  \\
    PoT Parallel & \textbf{2.88} & \textbf{2.68} & \textbf{2.04}	& \textbf{2.56}  \\
    PoT No Comment & 2.61 & \underline{2.41} & \underline{1.87} & \underline{2.28} \\
    \hline
  \end{tabular}
  }
  \caption{
  Code quality assessment with ICE-Score
  }
  \label{tab:code-quality}
\end{table}



\begin{figure}[htbp]
    \centering
    \includegraphics[width=\columnwidth, trim={0 0 0 0.6cm}, clip]{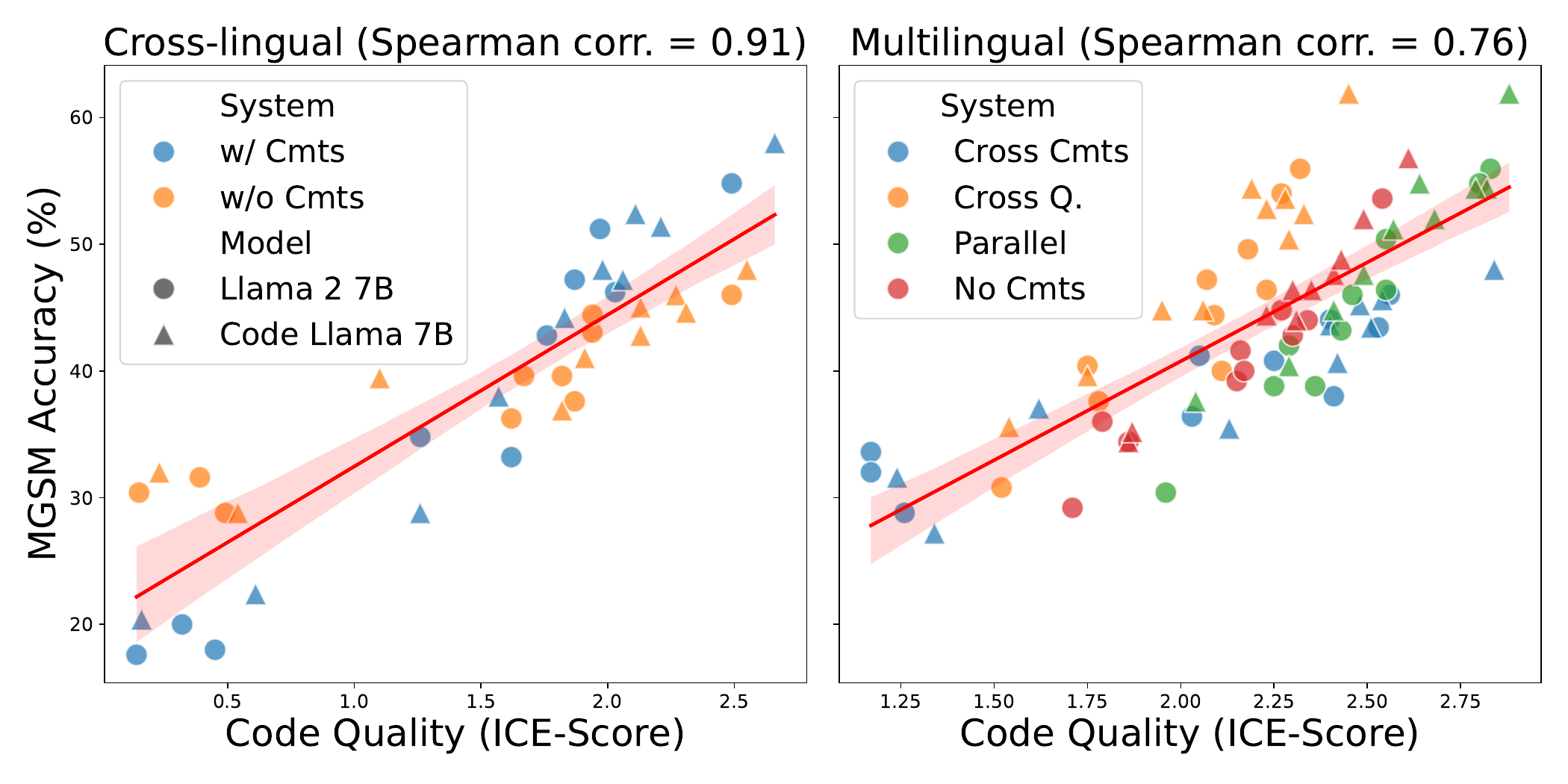}
    \caption{The relationship between code quality and answer accuracy in cross-lingual and multilingual settings. Each point represents a given language, considering a specific system and model combination.
    }
    \label{fig:sys-level-assoc}
\end{figure}

\textbf{System Level Correlation.}  Figure~\ref{fig:sys-level-assoc} illustrates a strong system-level correlation between MGSM accuracy and code quality, as measured by \texttt{ICE-Score}.
Across all finetuning strategies, in both cross-lingual and multilingual, we observe a consistent trend where higher code quality positively correlates with improved accuracy.
This relationship is quantified by a Spearman rank correlation coefficient of 0.91 and 0.76 for cross-lingual and multilingual, respectively.

Notably, this correlation persists across different model architectures and code generation conditions, reinforcing the importance of alignment strategies in enhancing both code quality and accuracy. 
These insights highlight the broader impact of alignment and resource availability on code generation, supporting the necessity of assessing the quality of intermediate outputs.

\textbf{Sample Level Association.}
Beyond system-level trends, we examined whether code quality can determine the correctness of individual solutions.
This relationship is demonstrated in Table \ref{tab:ice-score-distribution}, where the percentage distributions of \texttt{ICE-Score} for correct and incorrect answers show substantial differences across score ranges.
%
%
To further quantify this discriminative ability, we calculated the AUC for \texttt{ICE-Score} as a predictor of correctness, obtaining strong values of 0.94 and 0.96 for cross-lingual and multilingual settings, respectively.
Additionally, a t-test reveals a statistically significant difference between the correct and incorrect groups.
A detailed language-wise analysis is provided in Appendix~\ref{ap:code-analysis}.
\begin{table}[htbp]
  \resizebox{\columnwidth}{!}{
    \centering
    \caption{ICE-Score distribution (\%) for correct and incorrect answers in cross- and multilingual settings.}
    \label{tab:ice-score-distribution}
    \begin{tabular}{l l C{0.7cm} C{0.7cm} C{0.7cm} C{0.7cm} C{0.7cm}}
        \toprule
        \textbf{Setting} & \textbf{Answer Type} & \textbf{0} & \textbf{1} & \textbf{2} & \textbf{3} & \textbf{4} \\
        \midrule
        \multirow{2}{*}{Cross} & Correct   &  3.4 & 1.5 & 3.9 & 3.8 & 87.3  \\
                                       & Incorrect & 75.2  & 14.2  & 8.4 & 0.7 & 1.5 \\
        \midrule
        \multirow{2}{*}{Multi}  & Correct   & 2.0 & 1.3 & 3.6 & 4.1 & 89.0 \\
                                       & Incorrect & 52.8 & 25.8 & 17.5 & 2.3 & 1.6 \\
        \bottomrule
    \end{tabular}
  }
\end{table}

\textbf{Application in Test-Time Scaling.}
We now explore the potential of applying the ICE-Score as a heuristic for test-time scaling. 
We evaluate three approaches as discussed in Section~\ref{subsec:code_analysis}: (i) baseline model predictions without scaling, (ii) Self-Consistency (\texttt{SC}), and (iii) Soft Self-Consistency (\texttt{Soft-SC}) guided by the \texttt{ICE-Score}.
As shown in Table~\ref{tab:test-time-scaling}, our results indicate that test-time scaling substantially improves reasoning accuracy across both cross-lingual and multilingual settings. 
Conventional \texttt{SC} provides moderate gains, but \texttt{Soft-SC} with \texttt{ICE-Score} further boosts performance by prioritizing high-quality reasoning steps. 
Notably, for Llama2-7B, \texttt{Soft-SC} improves \textbf{cross-lingual performance from 39.2 to 56.6} and \textbf{multilingual performance from 57.2 to 71.2}. 
Similarly, CodeLlama-7B shows strong gains in both setups, demonstrating the method's robustness across model architectures.
These findings underscore the benefit of intermediate quality assessment as a means to improve cross-lingual and multilingual PoT reasoning and overall performance.

\begin{table}[htbp]
\tiny
  \centering
  \resizebox{\columnwidth}{!}{
  \begin{tabular}{l|c|cc|c}
    \hline
    \textbf{Method} & en & de & bn & ALL \\
    \hline\hline
    \multicolumn{5}{c}{\textit{Cross-lingual}} \\
    \hline    
    \multicolumn{1}{l|}{\underline{Llama2-7B}}& \multicolumn{1}{c|}{} & \multicolumn{2}{c|}{} & \multicolumn{1}{c}{} \\
    Without Comments & 58.0 & 40.4 & 12.0	&31.6  \\
    + \texttt{SC} & 65.2 & 51.6 & 15.2 &39.2  \\
    + \texttt{Soft-SC} (\texttt{ICE-Score}) & \textbf{76.8} & \textbf{69.2} & \textbf{33.6}	& \textbf{56.6}  \\ 
    \hline   
    \multicolumn{1}{l|}{\underline{CodeLlama-7B}}& \multicolumn{1}{c|}{} & \multicolumn{2}{c|}{} & \multicolumn{1}{c}{} \\
    Without Comments & 58.8 & 48.4 & 11.2 & 38.6  \\
    + \texttt{SC} & 69.6 & 57.2 & 17.2 & 46.7  \\
    + \texttt{Soft-SC} (\texttt{ICE-Score}) & \textbf{75.7} & \textbf{71.2} & \textbf{33.6} & \textbf{61.1}  \\
    \hline\hline
    \multicolumn{5}{c}{\textit{Multilingual}} \\
    \hline
    \multicolumn{1}{l|}{\underline{Llama2-7B}}& \multicolumn{1}{c|}{} & \multicolumn{2}{c|}{} & \multicolumn{1}{c}{} \\
    PoT Parallel & 56.0 & 47.2 & 30.8	& 44.6   \\
    + \texttt{SC} & 64.8 & 58.0 & 47.6	& 57.2  \\
    + \texttt{Soft-SC} (\texttt{ICE-Score}) & \textbf{77.6} & \textbf{72.0} & \textbf{65.6}	& \textbf{71.2} \\
    \hline   
    \multicolumn{1}{l|}{\underline{CodeLlama-7B}}& \multicolumn{1}{c|}{} & \multicolumn{2}{c|}{} & \multicolumn{1}{c}{} \\
    PoT Parallel & 61.9 & 52.8 & 35.6 & 49.0  \\
    + \texttt{SC} & 68.8 & 66.4 & 53.6 & 62.8  \\
    + \texttt{Soft-SC} (\texttt{ICE-Score}) & \textbf{79.2} & \textbf{77.6} & \textbf{68.8} & \textbf{75.6}  \\     
    \hline
  \end{tabular}
  }
  \caption{
  A comparative analysis of performance when implementing conventional \texttt{SC} and the proposed \texttt{Soft-SC} with \texttt{ICE-Score} in an optimal framework for cross-lingual and multilingual configurations.
  }
  \label{tab:test-time-scaling}
\end{table}

\section{Related Work}

\textbf{Mathematical Reasoning.}
Recent advancements in LLMs' mathematical reasoning capabilities have been driven by Chain-of-Thought (CoT) prompting \cite{og-cot-prompt, scratchpadNye}, which significantly outperforms direct-answer approaches by generating intermediate step-by-step reasoning.
Building on CoT, various enhancements have emerged, including self-consistency, which replaces greedy decoding with sampling-based inference to select the most consistent solution \cite{wang2023selfconsistency}.
Meanwhile, PoT and PaL \cite{pot, pal} improve reasoning by delegating computation to a Python interpreter, reducing the task of translating problems into code.

Another key advancement is instruction fine-tuning on mathematical datasets.
\citet{metamath} introduced MetaMathQA, expanding existing datasets through diverse rephrasings, while \citet{mammoth} leveraged a hybrid MathInstruct dataset combining CoT’s generality with PoT’s computational precision.
Additionally, external tool integration has been explored \cite{mario, tora}, with curated tool-use datasets enhancing LLMs’ reasoning capabilities.

\textbf{Multilingual Mathematical Reasoning.}
Despite LLMs' advancements in English mathematical reasoning, their performance in other languages still lags.
Efforts to bridge this gap include sample translation for multilingual alignment \cite{mathoctopus, mcot, qalign} and multilingual preference optimization \cite{mapo}.
\citet{mathoctopus} created a multilingual mathematical dataset by translating GSM8K into ten languages, though accurate translations remain a costly and time-consuming endeavor.
To mitigate this, \citet{qalign} proposed a two-step approach: translating questions into English before fine-tuning on larger English datasets like MetaMathQA.
Alternatively, \citet{mapo} leveraged existing translation models as alignment signals for preference optimization.

Beyond dataset translation, prompting techniques offer a cost-effective alternative.
\citet{not-all-lang-cross-lingual-cot-prompt} introduced role-playing prompts where the model first translates questions into English before applying CoT reasoning.
\citet{tot-multi-prompt} proposed a Tree-of-Thought framework for structured, multi-step reasoning across languages.
\citet{crosspal} extended PoT with Cross-PAL, aligning reasoning across multiple languages through code generation.

\section{Conclusion}

This study explores the effectiveness of Program-of-Thought (PoT) prompting for reasoning in cross- and multilingual settings by leveraging the reasoning-execution disentanglement concept.
We decompose the problem into two key challenges: (i) aligning multilingual questions with structured reasoning steps and (ii) assessing the impact of reasoning quality on final answer correctness.
Through systematic experimentation across cross-lingual and multilingual settings, we show that PoT fine-tuning substantially enhances reasoning alignment and generalization,
outperforming CoT fine-tuning.

Moreover, we establish a strong correlation between reasoning quality and answer accuracy. 
By leveraging \texttt{ICE-Score}-based inference strategies, we enhance performance,
particularly in low-resource languages. 
%
These findings provide insights into optimizing PoT for multilingual reasoning and open avenues for future research on improving reasoning alignment and execution.

Our findings contribute to a deeper understanding of multilingual PoT reasoning, providing insights into fine-tuning strategies and inference-time optimizations. 
Future work can extend this framework to additional reasoning-intensive tasks and explore more advanced alignment techniques to enhance PoT’s multilingual capabilities.

\section*{Limitations}

\textbf{GSM8K as a Reasoning Benchmark.}
The experimental setup of this study is grounded in grade school math problems from GSM8K; therefore, the results and key findings \emph{may not} generalize to other reasoning-intensive tasks. 
Furthermore, recent studies have raised concerns regarding potential data contamination~\cite{gsm-plus, gsm1k, mirzadeh2025gsmsymbolic}. 
Nonetheless, GSM8K remains the gold standard for assessing multi-step reasoning in the literature. 
We use this benchmark to ensure cross-comparability with existing work while emphasizing that our experimental framework is adaptable to any multi-step reasoning benchmark. 
In future work, we plan to extend our assessments to additional benchmarks to further validate our findings.

\textbf{ICE-Score Model Choice and Test-time Scaling.}
Our test-time scaling study presents a preliminary investigation into leveraging the ICE-Score as a Soft Self-Consistency (Soft-SC) heuristic. 
Prior work on ICE-Score~\cite{ice-score} suggests that stronger models yield better results.
This work prioritizes the evaluation accuracy of the code quality itself, verifying the correlation between the intermediate and end results. 
To this end, we employ the 405B variant of Llama3 for ICE-Score calculations in the correlation studies at the system and sample levels.
To maintain consistency, we continue to use this model for test-time scaling experiments.

Our findings indicate that incorporating ICE-Score into Soft-SC leads to performance improvements. 
However, the magnitude of these gains may depend on the specific ICE-Score model used. 
%
%
Future work should examine other ICE-Score configurations or alternative solutions, assessing their cost-benefit trade-off.



\section*{Acknowledgments}
This research is supported by the National Research Foundation, Singapore, under its National Large Language Models Funding Initiative. Any opinions, findings and conclusions or recommendations expressed in this material are those of the author(s) and do not reflect the views of National Research Foundation, Singapore


\bibliography{main}

\appendix

\section{MGSM8KInstruct}
\label{ap:mathoctopus}

We adopt MGSM8KInstruct~\cite{mathoctopus} as the reference dataset for CoT in multilingual settings.
This dataset comprises question-reasoning pairs $(\vb*{R}_i$, $\vb*{Q}_i)$ with $\vb*{Q}_i$ expressed in English, along with translations in nine additional languages, enabling the alignment of reasoning capabilities across different languages.
\citet{mathoctopus} introduced two training strategies:
\begin{inparaenum}[(i)]
    \item \emph{CoT Cross}: Incorporates English questions with answers in the target language, promoting multilingual adaptability. 
    Formally, the dataset is represented as: \begin{equation*}
        \mathcal{D}^\texttt{MGSM8KInstruct}_\texttt{cross} = \{(\vb*{Q}_i^\text{en}, \vb*{C}_i^l)|l\in L_\text{all}\}_{i=1}^N
    \label{eq:d-mathoctopus-cross}
    \end{equation*}
    where $L_\text{all}$ includes both English and target languages.
    \item \emph{CoT Parallel}: Uses question-answer pairs in the same language to 
    enhancing the PoT capability within each target language, denoted as: 
    \begin{equation*}
    \mathcal{D}^\texttt{MGSM8KInstruct}_\texttt{parallel} = \{(\vb*{Q}_i^l, \vb*{C}_i^l)|l\in L_\text{all}\}_{i=1}^N.
    \label{eq:d-mathoctopus-parallel}
    \end{equation*}
\end{inparaenum}

\section{PoT Generation Methods}
\label{ap:pot-syn}

  

To facilitate a fair comparison between PoT and CoT, we employ the GSM8K dataset, a collection of grade-school math problems that require 2-8 reasoning steps to solve, as the foundational benchmark.
As illustrated in Figure~\ref{fig:pipeline-cross}, we generate solutions in a programming language using an Oracle LLM through various methodologies:
\begin{compactenum}[1.]
    \item \textit{Zero-shot PoT Prompting}: Following the zero-shot prompting framework from \citet{pot}, the model is instructed to generate the \texttt{solver()} function in Python using a prompt $\vb*{S}_\text{PoT}$ with no exemplars. 
    Formally, the PoT synthesis from an Oracle LLM is represented as $\hat{\vb*{R}_i}\sim p_\text{Oracle}(\vb*{Q}_i|\vb*{S}_\text{PoT})$.
    
    \item \textit{Few-shot PoT Prompting}: Building on the methodologies of \citet{pot, pal}, $k$ in-context exemplars, $\vb*{E}_\text{FS}=\{(\vb*{Q}_1, \vb*{R}_1), ...,  (\vb*{Q}_k, \vb*{R}_k)\}$, are incorporated into the prompt to provide explicit guidance on desired outputs. 
    The PoT synthesis is thus defined as $\hat{\vb*{R}_i}\sim p_\text{Oracle}(\vb*{Q}_i|\vb*{E}_\text{FS},\vb*{S}_\text{PoT})$.
    
    \item \textit{Few-shot PoT Prompting + CoT Guidance}: Based on initial observations that high-quality PoT outputs often align with structured CoT reasoning $\vb*{C})$, an additional CoT guidance mechanism is introduced to better direct program generation. 
    In this setting, the examples $\vb*{E}_\text{FS-CoT}=\{(\vb*{Q}_1, \vb*{C}_1, \vb*{R}_1), ...,  (\vb*{Q}_k, \vb*{C}_k, \vb*{R}_k)\}$ include both CoT reasoning ($\vb*{C}_i$) and the corresponding PoT solution ($\vb*{R}_i$). 
    The PoT synthesis is then formulated as $\hat{\vb*{R}_i}\sim p_\text{Oracle}(\vb*{Q}_i|\vb*{C}_i, \vb*{E}_\text{FS-CoT},\vb*{S}_\text{PoT})$.
\end{compactenum}

%
We empirically tested three approaches to identify the most effective method for maximizing the match between program execution outputs and gold-standard answers, using Llama3.1 405B Instruct \cite{Llama3} as the Oracle LLM: zero-shot prompting, few-shot prompting, and few-shot prompting with CoT reasoning.
In zero-shot prompting, the model is given only the original GSM8K question and generates the corresponding Python code to solve it.
Few-shot prompting extends this by providing the model with two exemplars of correctly solved GSM8K questions along with their corresponding Python solutions. 
Few-shot prompting with CoT reasoning further builds upon this by incorporating both the original answer and its Chain-of-Thought (CoT) reasoning from GSM8K.
Our evaluation demonstrated that the few-shot + CoT approach consistently outperformed the other methods, achieving a correctness rate of 96.1\% in synthesizing PoT samples. In comparison, the few-shot prompting method yielded a correctness rate of 94.5\%, while the zero-shot approach resulted in a significantly lower accuracy of 58.7\%.

\begin{figure}[h]
    \small
    \centering
    \begin{mdframed}
    \textbf{System} \newline
    You are a helpful assistant. Answer the following question
    by implementing a solver() function in Python program
    step by step, and then return the answer.
    \newline
    Solve them in a step-by-step fashion and output a single option as the final answer in [language] language.
    \end{mdframed}
    \begin{mdframed}
    \textbf{User} \newline
    Question: [Question]
    \end{mdframed}    
    \caption{
    Zero-shot PoT prompt template for PoT synthesis, where [Question] serves as a placeholder for the problem statement.
    }
    \label{fig:zs-prompt}
\end{figure}

\begin{figure*}[h]
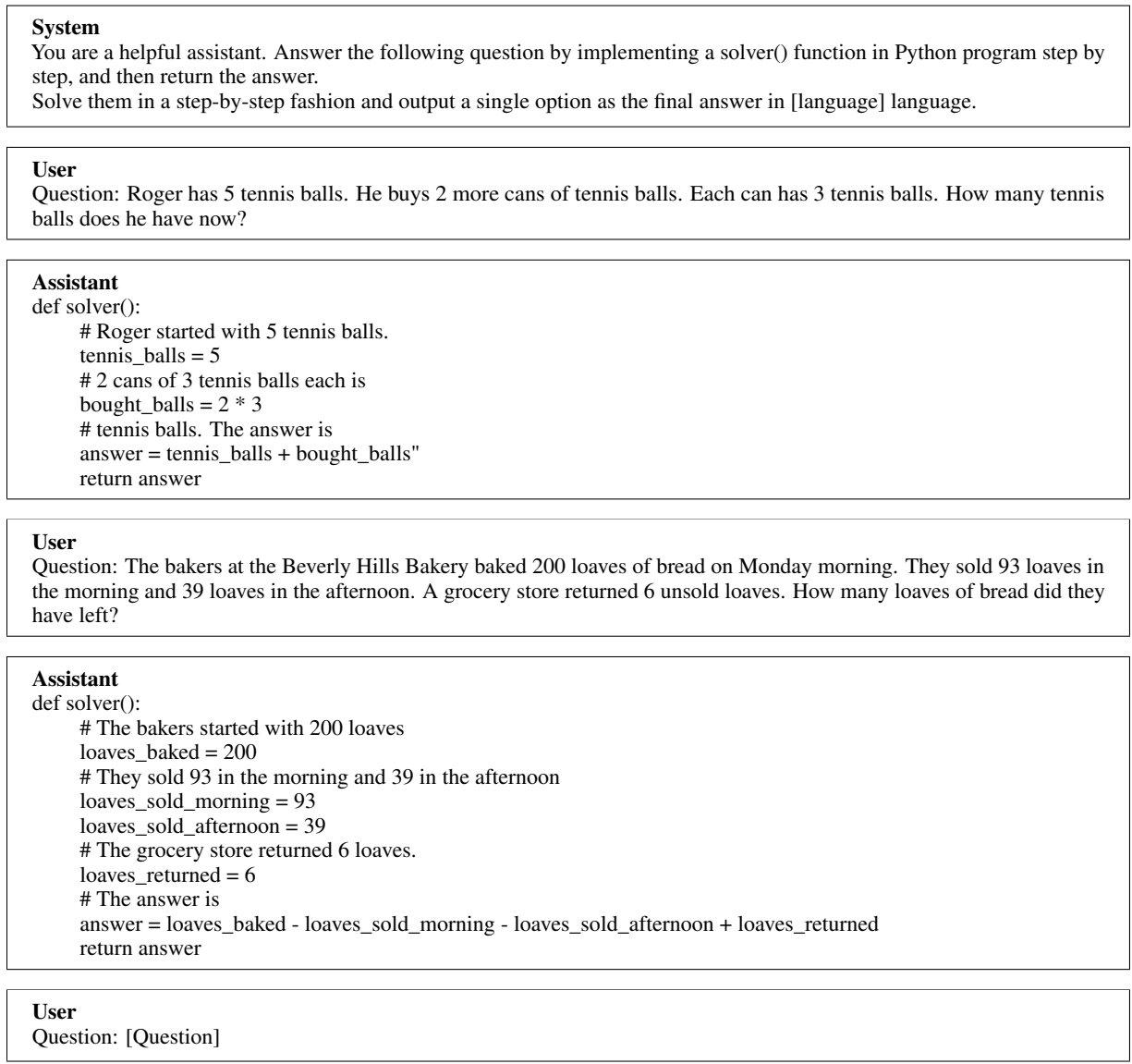

    \small
    \centering
    \begin{mdframed}
    \textbf{System} \newline
    You are a helpful assistant. Answer the following question
    by implementing a solver() function in Python program
    step by step, and then return the answer.
    \newline
    Solve them in a step-by-step fashion and output a single option as the final answer in [language] language.
    \end{mdframed}
    \begin{mdframed}
    \textbf{User} \newline
    Question: Roger has 5 tennis balls. He buys 2 more cans of tennis balls. Each can has 3 tennis balls. How many tennis balls does he have now?
    \end{mdframed}
    \begin{mdframed}
    \textbf{Assistant}\newline
    def solver():\newline\hspace*{6mm}    \# Roger started with 5 tennis balls.\newline\hspace*{6mm}    tennis\_balls = 5\newline\hspace*{6mm}    \# 2 cans of 3 tennis balls each is\newline\hspace*{6mm}    bought\_balls = 2 * 3\newline\hspace*{6mm}    \# tennis balls. The answer is\newline\hspace*{6mm}    answer = tennis\_balls + bought\_balls"\newline\hspace*{6mm}    return answer
    \end{mdframed}
    \begin{mdframed}
    \textbf{User}\newline
        Question: The bakers at the Beverly Hills Bakery baked 200 loaves of bread on Monday morning. They sold 93 loaves in the morning and 39 loaves in the afternoon. A grocery store returned 6 unsold loaves. How many loaves of bread did they have left?
    \end{mdframed}
    \begin{mdframed}
        \textbf{Assistant}\newline
        def solver():\newline\hspace*{6mm}    \# The bakers started with 200 loaves\newline\hspace*{6mm}    loaves\_baked = 200\newline\hspace*{6mm}    \# They sold 93 in the morning and 39 in the afternoon\newline\hspace*{6mm}    loaves\_sold\_morning = 93\newline\hspace*{6mm}    loaves\_sold\_afternoon = 39\newline\hspace*{6mm}    \# The grocery store returned 6 loaves.\newline\hspace*{6mm}    loaves\_returned = 6\newline\hspace*{6mm}    \# The answer is\newline\hspace*{6mm}    answer = loaves\_baked - loaves\_sold\_morning - loaves\_sold\_afternoon + loaves\_returned\newline\hspace*{6mm}    return answer
    \end{mdframed}
    \begin{mdframed}
        \textbf{User}\newline
        Question: [Question]
    \end{mdframed}
    \caption{
    Few-shot PoT prompt template for PoT synthesis, with exemplars adapted from \cite{pal, pot}.
    }
    \label{fig:fs-prompt}
\end{figure*}

\begin{figure*}[h]
    \small
    \centering
    \begin{mdframed}
    \textbf{System} \newline
    You are a helpful assistant. Answer the following question
    by implementing a solver() function in Python program
    step by step, and then return the answer.
    \newline
    Solve them in a step-by-step fashion and output a single option as the final answer in [language] language.
    \end{mdframed}
    \begin{mdframed}
    \textbf{User} \newline
    Question: Roger has 5 tennis balls. He buys 2 more cans of tennis balls. Each can has 3 tennis balls. How many tennis balls does he have now?\newline
    Chain-of-thought: Roger started with 5 tennis balls. 2 cans of 3 tennis balls each is 6 tennis balls. 5 + 6 = 11. The answer is 11.
    \end{mdframed}
    \begin{mdframed}
    \textbf{Assistant}\newline
    def solver():\newline\hspace*{6mm}    \# Roger started with 5 tennis balls.\newline\hspace*{6mm}    tennis\_balls = 5\newline\hspace*{6mm}    \# 2 cans of 3 tennis balls each is\newline\hspace*{6mm}    bought\_balls = 2 * 3\newline\hspace*{6mm}    \# tennis balls. The answer is\newline\hspace*{6mm}    answer = tennis\_balls + bought\_balls"\newline\hspace*{6mm}    return answer
    \end{mdframed}
    \begin{mdframed}
    \textbf{User}\newline
        Question: The bakers at the Beverly Hills Bakery baked 200 loaves of bread on Monday morning. They sold 93 loaves in the morning and 39 loaves in the afternoon. A grocery store returned 6 unsold loaves. How many loaves of bread did they have left?\newline
        Chain-of-thought: The bakers started with 200 loaves of bread. They sold 93 loaves in the morning and 39 loaves in the afternoon: 93 + 39 = 132 loaves sold. A grocery store returned 6 loaves, so they got 6 loaves back. 200 - 132 + 6 = 74 loaves left. The answer is 74.
    \end{mdframed}
    \begin{mdframed}
        \textbf{Assistant}\newline
        def solver():\newline\hspace*{6mm}    \# The bakers started with 200 loaves\newline\hspace*{6mm}    loaves\_baked = 200\newline\hspace*{6mm}    \# They sold 93 in the morning and 39 in the afternoon\newline\hspace*{6mm}    loaves\_sold\_morning = 93\newline\hspace*{6mm}    loaves\_sold\_afternoon = 39\newline\hspace*{6mm}    \# The grocery store returned 6 loaves.\newline\hspace*{6mm}    loaves\_returned = 6\newline\hspace*{6mm}    \# The answer is\newline\hspace*{6mm}    answer = loaves\_baked - loaves\_sold\_morning - loaves\_sold\_afternoon + loaves\_returned\newline\hspace*{6mm}    return answer
    \end{mdframed}
    \begin{mdframed}
        \textbf{User}\newline
        Question: [Question]\newline
        Chain-of-thought: [CoT]
    \end{mdframed}
    \caption{
    Few-shot PoT prompt template incorporating our proposed CoT-guided approach for PoT synthesis, where [CoT] serves as a placeholder for natural language reasoning.
    }
    \label{fig:fs-prompt}
\end{figure*}





\section{Training Setting}
Our code is primarily based on the MathOctopus codebase, with some minor modifications. The code will be made available.

\noindent\textbf{Prompt Template.} During training and testing, we consistently use the same prompt template from MathOctopus \cite{mathoctopus}.

\noindent\textbf{Setting.} We fully fintune all our models on a single 4xA100 node for three epochs with a maximum sequence length 1024.
For the Llama2 family and CodeLlama, we used a learning rate of 2e-5 and an effective batch size of 512.
However, we found that this setting caused the Llama3 8B model not to produce desirable results, which we discuss further in the next section.
Thus, we changed the effective batch size to 128 and the learning rate to 5e-6, following \cite{tulu3} for Llama 3 8B.
To generate multiple candidate predictions, we set \( top_k=50 \) and a temperature of 0.7, selecting the top 40 sequences for the voting process.

\section{Computing Resources}

We trained LLaMA family models on 4× NVIDIA A100 (80GB) GPUs, completing the fine-tuning process within approximately one hour for cross-lingual settings and around eight hours for multilingual settings.

During inference, generating predictions in a greedy fashion requires only three minutes. However, when producing multiple answer candidates with K=40, the process takes approximately seven hours to complete.

For Oracle LLM inference, we utilize a separate dedicated setup with 4× NVIDIA A100 (80GB) GPUs to host the LLM service, which is responsible for constructing PoT answers and evaluating code quality. The quality assessment process requires approximately 45 minutes for a single prediction and extends to 32 hours when assessing 40 candidates across all languages for a given model configuration. Additionally, we employ 62 concurrent processes to maximize inference throughput.

In summary, our experiments required a total of 544 A100 GPU hours for fine-tuning, 52 hours for inference, and 146 hours for quality assessment.

\section{Comparison with Non-Fine-Tuned PoT}

We compare our test-time scaling experiments with state-of-the-art (SOTA) non-fine-tuned PoT prompting methods and observe that our product models from PoT parallel with \texttt{SC} outperform SCross-PAL from \citet{crosspal} by 0.9 percentage points.
Furthermore, our proposed \texttt{Soft-SC} with \texttt{ICE-Score} achieves a significant accuracy improvement, increasing from 57.2\% to 71.2\%.

\begin{table}
\tiny
  \centering
  \resizebox{0.7\columnwidth}{!}{
  \begin{tabular}{l|c}
    \hline
    \textbf{Method} & ALL \\
    \hline\hline
    \multicolumn{2}{c}{\textit{Cross-lingual}} \\
    \hline    
    \multicolumn{1}{l|}{\underline{Llama2-7B}} & \multicolumn{1}{c}{} \\
    Without Comments &39.2  \\
    + \texttt{Soft-SC} (\texttt{ICE-Score}) & \textbf{56.6}  \\ 
    \hline   
    \multicolumn{1}{l|}{\underline{CodeLlama-7B}}& \multicolumn{1}{c}{} \\
    Without Comments & 38.6  \\
    + \texttt{SC} & 46.7  \\
    + \texttt{Soft-SC} (\texttt{ICE-Score}) & \textbf{61.1}  \\
    \hline\hline
    \multicolumn{2}{c}{\textit{Multilingual}} \\
    \hline
    \multicolumn{1}{l|}{\underline{Llama2-7B}}& \multicolumn{1}{c}{} \\
    PoT Parallel & 44.6   \\
    + \texttt{SC} & 57.2  \\
    + \texttt{Soft-SC} (\texttt{ICE-Score}) & \textbf{71.2} \\
    \hline   
    \multicolumn{1}{l|}{\underline{CodeLlama-7B}}& \multicolumn{1}{c}{} \\
    PoT Parallel & 49.0  \\
    + \texttt{SC} & 62.8  \\
    + \texttt{Soft-SC} (\texttt{ICE-Score}) & \textbf{75.6}  \\ 
    \hline\hline
    \multicolumn{2}{c}{\textit{Non-Fine-Tuned PoT}} \\
    \hline
    \multicolumn{1}{l|}{\underline{Llama2-7B}}& \multicolumn{1}{c}{} \\
    CLP \cite{clp} & 48.3 \\
    SCLP \cite{clp} & 54.1 \\
    \hdashline
    Cross-PAL \cite{crosspal} & 49.9 \\
    SCross-PAL \cite{crosspal} & \textbf{56.3} \\
    \hline
  \end{tabular}
  }
  \caption{
  The comparison of our adopted test-time scaling approaches with SOTA non-fine-tuned PoT approaches. The results of non-fune-tuned PoT are taken from \citet{crosspal}.
  }
  \label{tab:test-time-scaling-compare-prompt-pot}
\end{table}


\section{Sensitivity of Llama3-8B}
\begin{figure}[H]
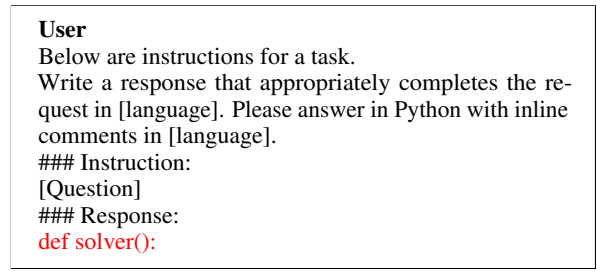

    \small
    \centering
    \begin{mdframed}
    \textbf{User} \newline
    Below are instructions for a task. \newline
    Write a response that appropriately completes the request in [language]. Please answer in Python with inline comments in [language].\newline
    \#\#\# Instruction: \newline[Question]\newline
    \#\#\# Response:\newline
    \textcolor{red}{def solver():} 
    \end{mdframed}
    \caption{
    Updated prompt with an added prefix (\texttt{def solver():}) for Llama3-8B.
    }
    \label{fig:llama3-prompt}
\end{figure}

During our testing, we observed that Llama3-8B exhibited significant sensitivity to our hyperparameters and chat template configurations.
Notably, the model frequently failed to generate the \texttt{def solver():} function header at the beginning of its reasoning chain, which is critical for extracting and compiling the generated code correctly.
To mitigate this issue, we inserted a prefix in our prompt, as illustrated in Figure \ref{fig:llama3-prompt}.
Additionally, with our initial hyperparameters, Llama3-8B frequently generated code snippets that failed to compile. Specifically, 9.12\% of its outputs were non-compilable, a significantly higher rate compared to Llama2-7B (3.08\%), CodeLlama-7B (2.04\%), and Llama2-13B (1.84\%). 
However, after refining our hyperparameters based on the approach outlined by \cite{tulu3} and adjusting the chat template, we observed a substantial reduction in compilation errors, with the failure rate dropping to 1.68\%.

\section{Alternative Metric For Code Quality Assessment}
Alternatively, to \texttt{ICE-Score}, we evaluated code quality using \texttt{CodeBERT-Score} \cite{codebertscore}. 
However, we noticed that GSM8K \cite{cobbe2021gsm8k} primarily consists of short code snippets where errors often involved small numerical mistakes rather than large structural or semantic differences.
Many of the errors stemmed from minor computation mistakes, like using the wrong arithmetic expression or associating wrong counts with the subject.
Since \texttt{CodeBERT-Score} is designed to assess broader semantic similarity, it struggled to distinguish the minute differences between correct and incorrect code.
As shown in Table \ref{tab:ablation-multi-codebert}, the scores across different systems varied only slightly $(\pm ~1.0\%)$, failing to reflect the accuracy differences observed in Tables \ref{tab:ablation-cross}, \ref{tab:ablation-multi}.
This suggests that \texttt{CodeBERT-Score} may not be well-suited for evaluating correctness in GSM8K-style problems.

\section{Full Tables}

\subsection{Main Results}
\label{ap:full-main}

This subsection serves as an extension of the results presented in Section~\ref{sec:QR-Alignment}.
In particular, we present the complete results for all fine-tuning alignment strategies in Table \ref{tab:ablation-cross} for cross-lingual settings and Table \ref{tab:ablation-multi} for multilingual settings.
These tables provide a detailed breakdown of performance across different configurations, reinforcing the trends observed in Section~\ref{sec:QR-Alignment}. 
The results confirm that PoT fine-tuning significantly improves multilingual reasoning, with cross-lingual generalization benefiting from the removal of inline comments and multilingual settings achieving higher alignment when comments are translated into target languages.

\begin{table*}[htbp]
\tiny
\centering
\resizebox{\textwidth}{!}{
\begin{tabular}{l|llllllllll|l}
\hline
\textbf{Method} & en & de & fr & es & ru & zh & ja & th & sw & bn & All \\
\hline
\multicolumn{1}{l|}{\underline{Llama2-7B}} & & & & & & & & & & & \\
With Comments & 58.3 & 37.9 & 40.0 & 44.4 & 39.2 & 33.2 & 25.1 & 6.8  & 5.2  & 9.9  & 30.0 \\
Without Comments & 58.0 & 40.4 & 40.4 & 43.6 & 37.1 & 38.4 & 32.7 & 7.6  & 5.6  & 12.0 & 31.6 \\
\hline
\multicolumn{1}{l|}{\underline{CodeLlama2-7B}} & & & & & & & & & & & \\
With Comments & 61.4 & 45.2 & 47.6 & 47.0 & 41.6 & 37.9 & 35.6 & 29.2 & 5.2  & 15.6 & 36.6 \\
Without Comments & 58.8 & 48.4 & 51.6 & 53.6 & 49.8 & 41.6 & 39.6 & 26.8 & 4.4  & 11.2 & 38.6 \\
\hline
\multicolumn{1}{l|}{\underline{Llama2-13B}} & & & & & & & & & & & \\
With Comments & 67.3 & 48.4 & 49.4 & 54.0 & 44.4 & 44.4 & 35.2 & 11.6 & 6.4  & 13.2 & 37.4 \\
Without Comments & 64.0 & 52.4 & 54.4 & 55.6 & 51.2 & 44.0 & 40.0 & 13.9 & 7.2  & 13.6 & 39.6 \\
\hline
\multicolumn{1}{l|}{\underline{Llama3-8B}} & & & & & & & & & & & \\
With Comments & 46.4 & 48.2 & 38.3 & 49.2  & 41.8 & 49.6  & 36.9  & 37.9 & 20.7 & 37.5 & 40.6 \\
Without Comments & 68.4  & 62.2 & 59.2  & 62.4  & 60.4  & 52.4  & 45.4 & 43.6  & 34.8  & 46.0   & 53.5 \\
\hline
\end{tabular}
}
\caption{
Accuracy (\%) on MGSM for all cross-lingual PoT variants, providing the full results corresponding to the subset shown in Table \ref{tab:pot-inline-comment}
}
\label{tab:ablation-cross}
\end{table*}

\begin{table*}[htbp]
\tiny
\centering
\resizebox{\textwidth}{!}{
\begin{tabular}{l|llllllllll|l}
\hline
\textbf{Method} & en & de & fr & es & ru & zh & ja & th & sw & bn & All \\
\hline
\multicolumn{1}{l|}{\underline{Llama2-7B}} & & & & & & & & & & & \\
PoT Cross Comment & 54.8 & 47.2 & 51.2 & 46.2 & 42.8 & 33.2 & 34.8 & 20.0 & 17.6 & 18.0 & 36.6 \\
PoT Cross Question & 46.0 & 37.6 & 43.0 & 44.4 & 39.6 & 39.6 & 36.3 & 31.6 & 30.4 & 28.8 & 37.7 \\
PoT Parallel & 56.0 & 47.2 & 46.4 & 54.0 & 49.6 & 44.4 & 40.0 & 40.4 & 37.6 & 30.8 & 44.6 \\
PoT No Comment & 53.6 & 41.6 & 42.8 & 44.8 & 44.0 & 39.2 & 40.0 & 36.0 & 34.4 & 29.2 & 40.6 \\
\hline
\multicolumn{1}{l|}{\underline{CodeLlama2-7B}} & & & & & & & & & & & \\
PoT Cross Comment & 58.0 & 47.2 & 51.4 & 52.4 & 48.0 & 44.2 & 38.0 & 28.8 & 20.4 & 22.4 & 41.1 \\
PoT Cross Question & 48.0 & 42.8 & 46.0 & 44.6 & 45.0 & 41.0 & 36.9 & 39.4 & 32.0 & 28.8 & 40.5 \\
PoT Parallel & 61.9 & 52.8 & 54.4 & 52.4 & 53.6 & 50.4 & 44.8 & 44.8 & 39.6 & 35.6 & 49.0 \\
PoT No Comment & 56.8 & 47.6 & 46.4 & 48.8 & 52.0 & 46.4 & 44.0 & 44.4 & 34.4 & 35.2 & 45.6 \\
\hline
\multicolumn{1}{l|}{\underline{Llama2-13B}} & & & & & & & & & & & \\
PoT Cross Comment & 62.0 & 53.6 & 52.4 & 54.8 & 50.0 & 42.0 & 39.2 & 21.6 & 23.2 & 23.2 & 42.2 \\
PoT Cross Question & 53.0 & 47.6 & 49.4 & 51.2 & 48.8 & 48.8 & 42.4 & 38.0 & 35.9 & 35.9 & 45.1 \\
PoT Parallel & 63.5 & 56.4 & 59.2 & 59.2 & 55.2 & 54.0 & 51.6 & 50.0 & 52.8 & 44.4 & 54.6 \\
PoT No Comment & 58.4 & 51.6 & 52.4 & 48.8 & 50.4 & 45.6 & 39.2 & 43.6 & 39.2 & 35.2 & 46.4 \\
\hline
\multicolumn{1}{l|}{\underline{Llama3-8B}} & & & & & & & & & & & \\
PoT Cross Comment & 72.8 & 62.4 & 66.4 & 67.2 & 63.6 & 52.0 & 49.6 & 52.0 & 46.2 & 51.2 & 58.3 \\
PoT Cross Question & 37.2 & 30.3 & 34.3 & 37.6 & 33.1 & 27.4 & 23.6 & 35.1 & 27.1 & 30.4 & 31.6\\
PoT Parallel & 76.5 & 64.4 & 63.2 & 66.4 & 64.0 & 63.2 & 56.4 & 57.6 & 59.6 & 55.2 & 62.6 \\
PoT No Comment & 65.2 & 60.0 & 59.6 & 59.2 & 57.2 & 57.2 & 49.4 & 55.2 & 53.6 & 48.4 & 56.5\\
\hline
\end{tabular}
}
\caption{
Accuracy (\%) on MGSM for all multilingual PoT variants, providing the full results corresponding to the subset shown in Table \ref{tab:multilingual-ablation} 
}
\label{tab:ablation-multi}
\end{table*}

\subsection{Code Analysis}
\label{ap:code-analysis}

This subsection extends the analysis presented in Section~\ref{section:code-analysis-results} by providing a full set of code quality evaluation results.
Our code analysis scores are presented in Table~\ref{tab:ablation-cross-icescore} and Table \ref{tab:ablation-multi-icescore} for cross-lingual and multilingual \texttt{ICE-Score}, respectively. Similarly, Table~\ref{tab:ablation-cross-codebert} and Table~\ref{tab:ablation-multi-codebert} provide the corresponding results for \texttt{CodeBERT-Score}.
These results are consistent with the findings in Section~\ref{section:code-analysis-results}, confirming a strong correlation between reasoning quality and final answer correctness. 
The observed trends support the effectiveness of leveraging code quality for test-time scaling, with improvements in underrepresented languages being particularly notable.

\begin{table*}[htbp]
\tiny
\centering
\resizebox{\textwidth}{!}{
\begin{tabular}{l|llllllllll|l}
\hline
\textbf{Method} & en & de & fr & es & ru & zh & ja & th & sw & bn & All \\
\hline
\multicolumn{1}{l|}{\underline{Llama2-7B}} & & & & & & & & & & & \\
With Comments & 2.40 & 1.87 & 1.84 & 2.03 & 1.76 & 1.41 & 1.26 & 0.32 & 0.14 & 0.45 & 1.35 \\
Without Comments & 2.49 & 1.87 & 1.94 & 1.94 & 1.82 & 1.67 & 1.62 & 0.39 & 0.15 & 0.49 & 1.44 \\
\hline
\multicolumn{1}{l|}{\underline{CodeLlama2-7B}} & & & & & & & & & & & \\
With Comments & 2.66 & 2.06 & 2.21 & 2.11 & 1.98 & 1.83 & 1.57 & 1.26 & 0.16 & 0.61 & 1.65 \\
Without Comments & 2.54 & 2.13 & 2.15 & 2.31 & 2.13 & 1.85 & 1.82 & 1.10 & 0.23 & 0.54 & 1.68 \\
\hline
\multicolumn{1}{l|}{\underline{Llama2-13B}} & & & & & & & & & & & \\
With Comments & 2.79 & 2.21 & 2.29 & 2.37 & 2.02 & 2.04 & 1.76 & 0.56 & 0.21 & 0.60 & 1.69 \\
Without Comments & 2.49 & 1.87 & 1.94 & 1.94 & 1.82 & 1.67 & 1.62 & 0.39 & 0.15 & 0.49 & 1.44 \\
\hline
\multicolumn{1}{l|}{\underline{Llama3-8B}} & & & & & & & & & & & \\
With Comments & 2.74 & 2.32 & 2.40 & 2.63 & 2.36 & 2.07 & 1.86 & 2.20 & 1.34 & 1.81 & 2.17 \\
Without Comments & 2.75 & 2.46 & 2.61 & 2.56 & 2.54 & 2.00 & 1.92 & 1.88 & 1.64 & 2.00 & 2.24 \\
\hline
\end{tabular}
}
\caption{
\texttt{ICE-Score} on MGSM for all cross-lingual PoT variants, providing the full results corresponding to the subset shown in Table \ref{tab:code-quality} 
}
\label{tab:ablation-cross-icescore}
\end{table*}

\begin{table*}[htbp]
\tiny
\centering
\resizebox{\textwidth}{!}{
\begin{tabular}{l|llllllllll|l}
\hline
\textbf{Method} & en & de & fr & es & ru & zh & ja & th & sw & bn & All \\
\hline
\multicolumn{1}{l|}{\underline{Llama2-7B}} & & & & & & & & & & & \\
PoT Cross Comment & 2.56 & 2.41 & 2.53 & 2.40 & 2.25 & 2.05	& 2.03 & 1.17 & 1.17 & 1.26 & 1.98\\
PoT Cross Question & 2.32 & 2.07 & 2.23 & 2.27 & 2.18 & 2.09 & 2.11 & 1.75 & 1.78 & 1.52 & 2.03\\
PoT Parallel & 2.83 & 2.55 & 2.46 & 2.80 & 2.55 & 2.43 & 2.25 & 2.29 & 2.36 & 1.96 & 2.45\\
PoT No Comment & 2.54 & 2.16 & 2.30 & 2.27 & 2.34 & 2.15 & 2.17 & 1.79 & 1.86 & 1.71 & 2.13\\
\hline
\multicolumn{1}{l|}{\underline{CodeLlama2-7B}} & & & & & & & & & & & \\
PoT Cross Comment & 2.84 & 2.40 & 2.54 & 2.48 & 2.51 & 2.42 & 2.13 & 1.62 & 1.24 & 1.34 & 2.15\\
PoT Cross Question & 2.45 & 2.23 & 2.19 & 2.33 & 2.28 & 2.29 & 2.06 & 1.95 & 1.75 & 1.54 & 2.11\\
PoT Parallel & 2.88 & 2.68 & 2.64 & 2.82 & 2.79 & 2.57 & 2.49 & 2.41 & 2.29 & 2.04 & 2.56\\
PoT No Comment & 2.61 & 2.41 & 2.35 & 2.43 & 2.49 & 2.30 & 2.31 & 2.23 & 1.86 & 1.87 & 2.28\\
\hline
\multicolumn{1}{l|}{\underline{Llama2-13B}} & & & & & & & & & & & \\
PoT Cross Comment & 2.91 & 2.76 & 2.67 & 2.88 & 2.58 & 2.50 & 2.25 & 1.40 & 1.65 & 1.55 & 2.31\\
PoT Cross Question & 2.69 & 2.48 & 2.48 & 2.58 & 2.52 & 2.48 & 2.30 & 2.12 & 2.12 & 1.84 & 2.36\\
PoT Parallel & 2.94 & 2.91 & 2.90 & 2.82 & 2.93 & 2.81 & 2.79 & 2.70 & 2.75 & 2.36 & 2.79\\
PoT No Comment & 2.70 & 2.52 & 2.62 & 2.55 & 2.45 & 2.45 & 2.30 & 2.30 & 2.26 & 2.17 & 2.43\\
\hline
\multicolumn{1}{l|}{\underline{Llama3-8B}} & & & & & & & & & & & \\
PoT Cross Comment & 3.12 & 2.75 & 2.87 & 2.90 & 2.84 & 2.44 & 2.42 & 2.32 & 2.13 & 2.38 & 2.61 \\
PoT Cross Question & 2.51 & 2.07 & 2.31 & 2.25 & 2.13 & 2.04 & 1.95 & 2.11 & 1.67 & 1.83 & 2.09\\
PoT Parallel & 3.15 & 2.88 & 2.82 & 3.03 & 2.84 & 2.82 & 2.58 & 2.72 & 2.68 & 2.56 & 2.81 \\
PoT No Comment & 2.76 & 2.54 & 2.47 & 2.53 & 2.48 & 2.56 & 2.24 & 2.46 & 2.33 & 2.10 & 2.45 \\
\hline
\end{tabular}
}
\caption{
\texttt{ICE-Score} on MGSM for all multilingual PoT variants, providing the full results corresponding to the subset shown in Table \ref{tab:code-quality}  
}
\label{tab:ablation-multi-icescore}
\end{table*}

\begin{table*}[htbp]
\tiny
\centering
\resizebox{\textwidth}{!}{
\begin{tabular}{l|llllllllll|l}
\hline
\textbf{Method} & en & de & fr & es & ru & zh & ja & th & sw & bn & All \\
\hline
\multicolumn{1}{l|}{\underline{Llama2-7B}} & & & & & & & & & & & \\
With Comments & 90.50 & 83.53 & 84.18 & 83.21 & 85.68 & 85.94 & 86.02 & 81.33 & 79.24 & 81.81 & 84.14 \\
Without Comments & 90.08 & 81.56 & 82.03 & 81.16 & 83.94 & 85.66 & 86.61 & 82.15 & 78.56 & 81.92 & 83.37 \\
\hline
\multicolumn{1}{l|}{\underline{CodeLlama2-7B}} & & & & & & & & & & & \\
With Comments & 90.69 & 85.23 & 84.00 & 82.80 & 86.33 & 86.15 & 86.40 & 84.14 & 78.87 & 83.04 & 84.77 \\
Without Comments & 89.85 & 84.67 & 83.88 & 83.11 & 86.54 & 85.68 & 86.74 & 84.07 & 79.61 & 83.21 & 84.74 \\
\hline
\multicolumn{1}{l|}{\underline{Llama2-13B}} & & & & & & & & & & & \\
With Comments & 90.50 & 84.66 & 84.29 & 83.75 & 85.12 & 87.10 & 87.05 & 82.00 & 80.71 & 82.82 & 84.80 \\
Without Comments & 90.29 & 85.62 & 85.08 & 84.78 & 86.72 & 87.06 & 87.64 & 83.52 & 81.53 & 83.60 & 85.58 \\
\hline
\multicolumn{1}{l|}{\underline{Llama3-8B}} & & & & & & & & & & & \\
With Comments & 83.23 & 80.30 & 79.90 & 81.22 & 80.36 & 80.06 & 79.58 & 80.65 & 78.61 & 79.64 & 80.36 \\
Without Comments & 84.36 & 82.11 & 81.60 & 80.38 & 82.23 & 79.26 & 78.16 & 79.95 & 78.34 & 79.50 & 80.59 \\
\hline
\end{tabular}
}
\caption{
\texttt{CodeBERT-Score} (F1) on MGSM for all cross-lingual PoT variants. 
}
\label{tab:ablation-cross-codebert}
\end{table*}

\begin{table*}[htbp]
\small
\centering
\resizebox{\textwidth}{!}{
\begin{tabular}{l|llllllllll|l}
\hline
\textbf{Method} & en & de & fr & es & ru & zh & ja & th & sw & bn & All \\
\hline
\multicolumn{1}{l|}{\underline{Llama2-7B}} & & & & & & & & & & & \\
PoT Cross Comment & 90.26 & 88.07 & 87.94 & 87.56 & 87.38 & 86.81 & 87.13 & 85.03 & 82.63 & 84.33 & 86.71\\
PoT Cross Question & 89.13 & 88.41 & 88.12 & 88.16 & 88.39 & 87.22 & 87.60 & 86.11 & 86.24 & 86.18 & 87.56\\
PoT Parallel & 89.95 & 89.14 & 88.90 & 89.27 & 88.72 & 87.99 & 88.34 & 87.35 & 87.53 & 86.89 & 88.41\\
PoT No Comment & 89.73 & 88.74 & 88.29 & 88.60 & 88.13 & 87.24 & 88.14 & 86.64 & 86.79 & 86.62 & 87.89\\
\hline
\multicolumn{1}{l|}{\underline{CodeLlama2-7B}} & & & & & & & & & & & \\
PoT Cross Comment & 90.95 & 88.38 & 88.45 & 88.74 & 87.89 & 87.75 & 88.23 & 84.79 & 83.55 & 85.23 & 87.40\\
PoT Cross Question & 89.10 & 88.26 & 87.84 & 87.79 & 87.25 & 87.06 & 87.03 & 86.45 & 86.02 &85.95 & 87.28\\
PoT Parallel & 90.08 & 88.79 & 89.03 & 88.77 & 88.47 & 87.69 & 87.68 & 87.37 & 86.55 & 87.09 & 88.15\\
PoT No Comment & 89.84 & 88.44 & 88.71 & 88.66 & 88.04 & 87.47 & 87.49 & 86.90 & 86.55 & 86.14 & 87.82\\
\hline
\multicolumn{1}{l|}{\underline{Llama2-13B}} & & & & & & & & & & & \\
PoT Cross Comment & 90.26 & 87.80 & 87.75 & 87.36 & 87.27 & 86.83 & 87.53 & 83.72 & 82.91 & 84.05 & 86.55\\
PoT Cross Question & 89.92 & 88.84 & 88.63 & 88.65 & 88.48 & 87.97 & 88.11 & 86.92 & 87.21 & 87.13 & 88.19\\
PoT Parallel & 90.40 & 89.53 & 89.32 & 89.51 & 89.55 & 88.53 & 88.92 & 88.10 & 87.93 & 87.57 & 88.94\\
PoT No Comment & 89.99 & 89.02 & 89.14 & 89.00 & 88.74 & 88.40 & 88.44 & 87.30 & 87.10 & 87.04 & 88.42\\
\hline
\multicolumn{1}{l|}{\underline{Llama3-8B}} & & & & & & & & & & & \\
PoT Cross Comment & 91.02 & 89.32 & 88.69 & 89.06 & 89.01 & 88.29 & 88.32 & 87.91 & 86.59 & 87.55 & 88.58\\
PoT Cross Question & 80.30 & 78.94 & 78.95 & 79.61 & 79.63 & 79.22 & 79.51 & 79.18 & 79.12 & 79.63 & 79.38\\
PoT Parallel & 90.49 & 90.17 & 89.81 & 89.31 & 89.38 & 88.71 & 88.99 & 88.90 & 88.40 & 88.75 & 89.29\\
PoT No Comment & 89.80 & 88.81 & 88.78 & 89.07 & 88.40 & 88.35 & 88.70 & 88.37 & 88.08 & 88.10 & 88.65\\
\hline
\end{tabular}
}
\caption{
\texttt{CodeBERT-Score} (F1) on MGSM in for all multilingual PoT variants. 
}
\label{tab:ablation-multi-codebert}
\end{table*}

\begin{table*}[htbp]
\small
\centering
\resizebox{\textwidth}{!}{%
\begin{tabular}{l|llllllllll}
\hline
Method & en & de & fr & es & ru & zh & ja & th & sw & bn \\
\hline \hline
\multicolumn{11}{c}{\textit{Cross-lingual}} \\
\hline    
\underline{Llama2-7B} & & & & & & & & & & \\ 
AUC  & 0.9659  & 0.9544  & 0.9782  & 0.9728  & 0.9733  & 0.9622  & 0.9680  & 0.9683  & 0.7376  & 0.9068  \\
T-Statsitic     & 28.66   & 25.67   & 32.72   & 31.29   & 31.82   & 28.81   & 29.04   & 12.87   & 3.23    & 10.56   \\
\hline   
\underline{CodeLlama-7B} & & & & & & & & & & \\ 
AUC  & 0.9752  & 0.9708  & 0.9661  & 0.9736  & 0.9723  & 0.9606  & 0.9528  & 0.9215  & 0.8524  & 0.9096  \\
T-Statsitic     & 28.85   & 29.85   & 29.82   & 31.68   & 31.36   & 28.86   & 25.30   & 16.48   & 4.96    & 9.96    \\
\hline \hline
\multicolumn{11}{c}{\textit{Multilingual}} \\
\hline
\underline{Llama2-7B} & & & & & & & & & & \\ 
AUC  & 0.9847  & 0.9561  & 0.9725  & 0.9771  & 0.9675  & 0.9714  & 0.9865  & 0.9556  & 0.9437  & 0.9480  \\
T-Statsitic     & 31.21   & 25.70   & 28.94   & 28.51   & 29.04   & 27.80   & 36.85   & 25.14   & 22.99   & 21.79   \\
\hline   
\underline{CodeLlama-7B} & & & & & & & & & & \\ 
AUC  & 0.9627  & 0.9327  & 0.9627  & 0.9756  & 0.9536  & 0.9511  & 0.9476  & 0.9532  & 0.9537  & 0.9476  \\
T-Statsitic     & 25.13   & 21.46   & 25.94   & 28.20   & 23.93   & 23.45   & 22.96   & 25.07   & 24.10   & 23.96   \\
\hline
\end{tabular}
}
\captionsetup{justification=centerlast}
\caption{T-Statistic and AUC scores for Llama2-7B and CodeLlama-7B across MGSM}
\label{tab:ttest-stats}
\end{table*}

\end{document}